\documentclass[journal]{IEEEtran}
\usepackage{cite}
\usepackage{verbatim}

\ifCLASSINFOpdf
\usepackage[pdftex]{graphicx}
\usepackage{graphicx}
\usepackage{float}
\usepackage{booktabs}
\usepackage{multirow}
\usepackage{amssymb}
\usepackage{booktabs}
\usepackage{subfigure}
\usepackage{caption}
\else
\fi
\usepackage{amsmath}
\usepackage{mathrsfs}
\usepackage{amsfonts}
\usepackage{bm}
\allowdisplaybreaks[4]

\usepackage{stfloats}
\usepackage{enumerate}
\usepackage{enumitem}
\begin{document}
%
\title{Stripe-based and Attribute-aware Network: A Two-Branch Deep Model for Vehicle Re-identification}
%
%
%

\author{Jingjing~Qian,~
        Wei~Jiang,~
        Hao~Luo,~
        and~Hongyan~Yu
\thanks{J. Qian, W. Jiang, and H. Luo are with the College of Control Science and Engineering, Zhejiang University, Hangzhou 310007, China (e-mail: jingjingqian@zju.edu.cn;  jiangwei\_zju@zju.edu.cn; haoluocsc@zju.edu.cn).}
\thanks{H. Yu is with the Beijing Electro-mechanical Engineering Institute, Beijing 100074, China (e-mail: yuhongyan09@163.com).}
\thanks{Manuscript received October 12, 2019. This work was supported by the National Natural Science Foundation of China under Grant 61633019 and the Science Foundation of Chinese Aerospace Industry under Grant JCKY2018204B053. \emph{(Corresponding author: Wei Jiang.)}}}

\maketitle

\begin{abstract}
Vehicle re-identification (Re-ID) has been attracting increasing interest in the field of computer vision due to the growing utilization of surveillance cameras in public security. However, vehicle Re-ID still suffers a similarity challenge despite the efforts made to solve this problem. This challenge involves distinguishing different instances with nearly identical appearances. In this paper, we propose a novel two-branch stripe-based and attribute-aware deep convolutional neural network (SAN) to learn the efficient feature embedding for vehicle Re-ID task. The two-branch neural network, consisting of stripe-based branch and attribute-aware branches, can adaptively extract the discriminative features from the visual appearance of vehicles. A horizontal average pooling and dimension-reduced convolutional layers are inserted into the stripe-based branch to achieve part-level features. Meanwhile, the attribute-aware branch extracts the global feature under the supervision of vehicle attribute labels to separate the similar vehicle identities with different attribute annotations. Finally, the part-level and global features are concatenated together to form the final descriptor of the input image for vehicle Re-ID. The final descriptor not only can separate vehicles with different attributes but also distinguish vehicle identities with the same attributes. The extensive experiments on both VehicleID and VeRi databases show that the proposed SAN method outperforms other state-of-the-art vehicle Re-ID approaches.
\end{abstract}


%
\IEEEpeerreviewmaketitle

\section{Introduction}
\IEEEPARstart{A}{t} present, vehicle search and re-identification (Re-ID) are receiving increasing research interest in the field of computer vision due to their important applications in video surveillance. Specifically, vehicle Re-ID is the problem of identifying the same vehicle across different surveillance camera views. In practical scenarios, vehicle Re-ID is a very challenging computer vision problem, due to the inconspicuous divergences among different instances. Therefore, how to develop an effective vehicle Re-ID method has attracted more and more attention.
\par
\begin{figure}
\centering
\subfigure[]{
\begin{minipage}[b]{0.17\linewidth}
\includegraphics[width=1\linewidth]{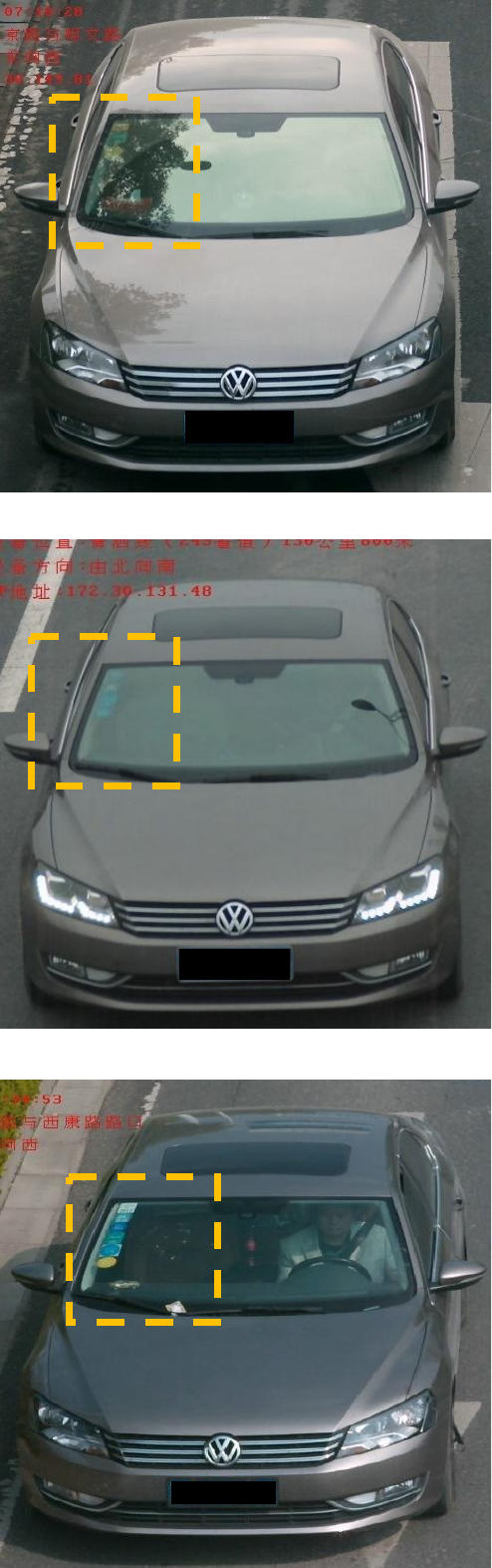}
\end{minipage}}
\subfigure[]{
\begin{minipage}[b]{0.17\linewidth}
\includegraphics[width=1\linewidth]{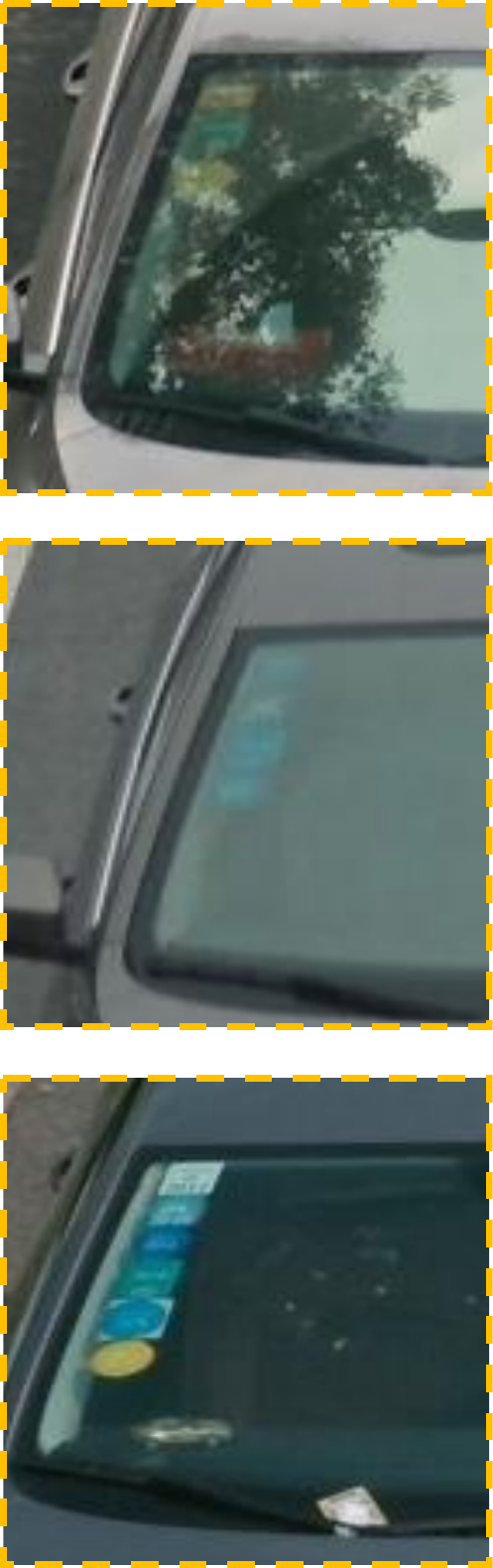}
\end{minipage}}
\subfigure[]{
\begin{minipage}[b]{0.17\linewidth}
\includegraphics[width=1\linewidth]{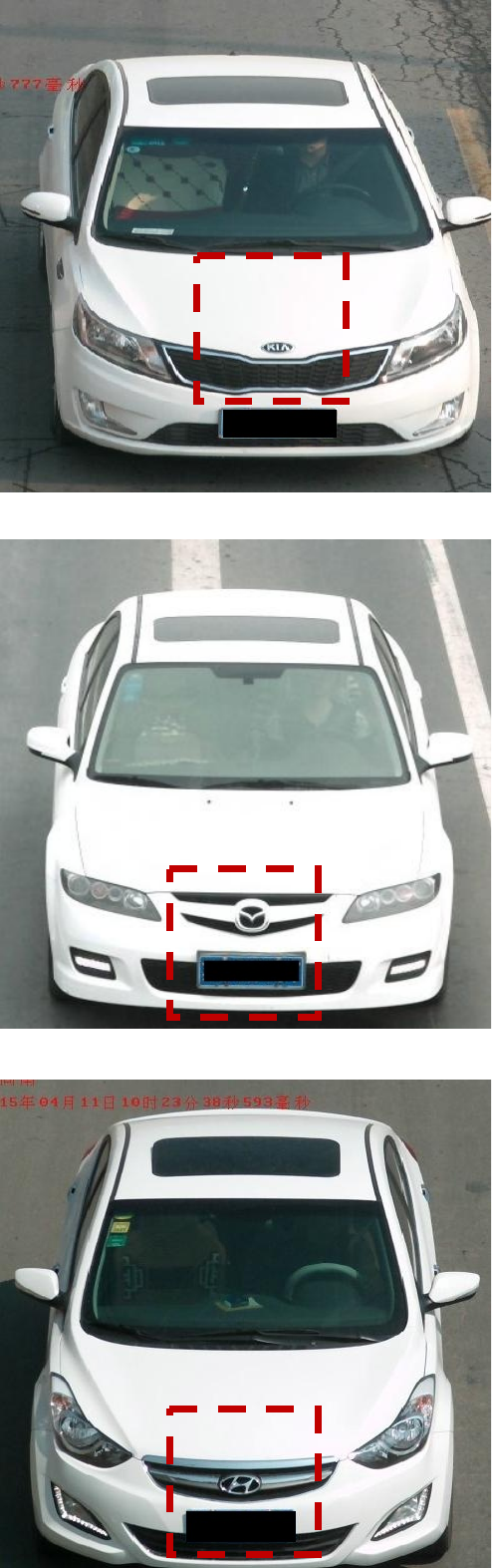}
\end{minipage}}
\subfigure[]{
\begin{minipage}[b]{0.17\linewidth}
\includegraphics[width=1.01\linewidth]{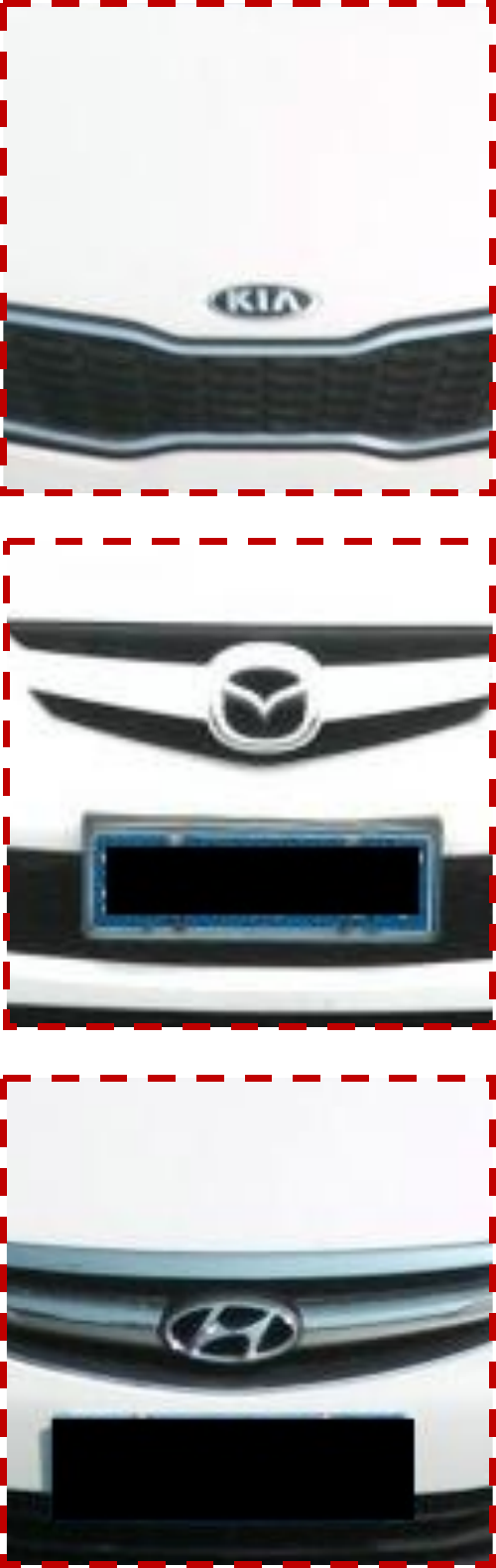}
\end{minipage}}
\subfigure[]{
\begin{minipage}[b]{0.17\linewidth}
\includegraphics[width=1\linewidth]{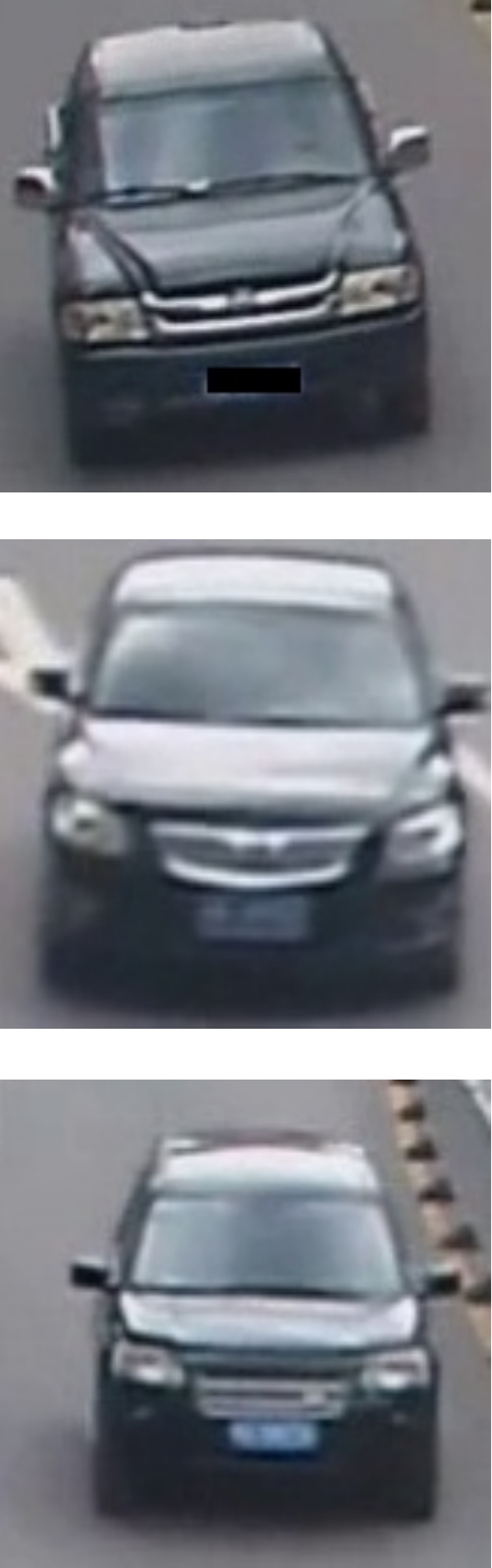}
\end{minipage}}
\caption{(a) Three vehicle images come from different vehicle identities with similar appearance. Apparently, distinguishing them is difficult because all three vehicle identities come from the same vehicle model. (b) Discriminative vehicle parts of images in (a). (c) Three vehicle images come from different vehicle identities with similar appearance. Although they belong to different models, distinguishing them is difficult. (d) Discriminative vehicle logos of images in (c). (e) Three vehicle images of VeRi database come from different vehicle identities with similar appearance. Although they have different types, identifying them is difficult.}
\label{intro}
\end{figure}
Despite serving as a unique ID of a vehicle, license plates are not always reliable because they can be easily faked. In addition, plates are sometimes unrecognizable because of the low resolution of the license image. Therefore, appearance-based vehicle Re-ID plays an important role in real-world applications.
To address the problem of appearance based vehicle Re-ID, two large benchmark databases, namely, VehicleID \cite{liu2016deepdeep} and VeRi \cite{liu2016deep} were released. On the basis of these databases, several deep learning-based vehicle Re-ID methods \cite{liu2016deep, liu2016large, zapletal2016vehicle} have been proposed to develop an efficient global feature representation for each vehicle, while researchers in \cite{liu2016deepdeep, yuan2017hard, yan2017exploiting, xu2017learning} used a distance metric to pull matched image pairs closer and mismatched ones farther. However, both of these two strategies are flawed in some aspects. First, although the deep learning-based networks have the ability to learn global semantic features from an entire vehicle image, these networks still fail to focus on some small but discriminative regions, such as inspection marks on windscreen and personalized decorations inside, as shown in  Fig. \ref {intro}. Such discriminative regions are important in identifying a vehicle when different vehicles share an extremely similar appearance. Moreover, the metric learning methods treat images in mismatched pairs equally and ignore the hierarchical multigrain relationships among vehicles.
\par
In this paper, we explore the similarity phenomenon in vehicle Re-ID. As illustrated in Fig. \ref {intro} (a), different vehicles usually share similar geometric shapes and appearances which can be difficult to distinguish for deep models. However, variances are present in part-level features (e.g., stickers in windows), which are easily recognizable by human beings, see Fig. \ref {intro} (b). Moreover, different vehicles can be very similar despite belonging to different model, as shown in Fig. \ref {intro} (c). Therefore model annotations are obviously beneficial for vehicle Re-ID, see Fig. \ref {intro} (d). In our paper, two vehicle databases, namely, VehicleID and VeRi are utilized for the experiments. The train set of VehicleID database has the model labels, whereas the VeRi database not. Fortunately, the type information is annotated in VeRi database. As illustrated in Fig. \ref {intro} (e), the vehicle images in the VeRi database are extremely similar despite belonging under different types. As a result, type annotations are also helpful for vehicle Re-ID. 
\par
To handle the similarity phenomenon in vehicle Re-ID, we propose a two-branch stripe-based and attribute-aware deep convolutional neural network (SAN) for vehicle re-identification, which integrates the part-level features and attribute information into a unified architecture. The proposed SAN has two branches, namely, stripe-based branch and attribute-aware branch. The former is developed to achieve discriminative part-level features, which are important in distinguishing different vehicle identities. On the other hand, the latter utilizes attribute labels to separate the similar vehicle identities with different attribute annotations. Specifically, in the attribute-aware branch, the model and type labels of the VehicleID and VeRi databases are adopted, respectively.
\par
In summary, the contribution of our work is three-folds:
\begin{enumerate}
\item
An effective representation learning framework is designed by jointly considering part-level and global representations.
\item
A novel two-branch SAN is proposed, in which part-level features and attribute information are combined to enhance the discriminative capability of the final descriptor for vehicle Re-ID.
\item
State-of-the-art results are achieved in the two challenging datasets: Rank-1 accuracy of 79.7\%, 78.4\%, and 75.6\% in the three test sets of VehicleID; and mean average precision (mAP) and Rank-1 accuracies of 72.5\% and 93.3\%, respectively, on the test set of VeRi.
\end{enumerate}
\par
The rest of this paper is organized as follows. Section \uppercase\expandafter{\romannumeral2} introduces the related works. Section \uppercase\expandafter{\romannumeral3} describes the proposed two-branch SAN for vehicle Re-ID. Section \uppercase\expandafter{\romannumeral4} presents the experimental results that validate the superiority of the proposed method. Section \uppercase\expandafter{\romannumeral5} concludes this paper.
\section{Related Works}
\subsection{Re-identification (Re-ID)}
Re-identification (Re-ID) is widely studied in the field of computer vision. This task possesses various important applications. Most existing Re-ID methods focus on person Re-ID problems, which aim to determine the target persons in a large gallery set using probe images. Recently, CNN-based features have achieved great progress on person Re-ID. 
Authors in \cite{cheng2016person, zhu2017part} utilized local cues by extracting multiple patches from a loosely associated image of human body parts. In addition, some attempts \cite{schumann2017person, su2017multi} have been made  to improve person Re-ID performance using person attributes. Almazan \emph{et al.}\cite{almazan2018re} introduced a list of good practices that can design and train an efficient image representation model for person Re-ID. The key practices include hard triplet mining, pretraining for identity classification, dataset augmentation with difficult examples, sufficiently large image resolution, and state-of-the-art base architecture. Triplet mining and ID classification have been already adopted in vehicle Re-ID \cite{bai2018group, kumar2019vehicle}. Kumar \emph{et al.}\cite{kumar2019vehicle} demonstrated an extensive evaluation of contrastive and triplet losses during vehicle Re-ID.
\subsection{Vehicle Re-ID}
Recently, vehicle Re-ID has gained increasing attention from scholars in related fields. Liu \emph{et al.}\cite{liu2016deepdeep} proposed a new large-scale vehicle Re-ID dataset VehicleID, which contains 26,267 different vehicles with over 222,000 images collected from real surveillance cameras and labeled at the identity level. They also introduced a pipeline that uses deep relative distance learning (DRDL) to project vehicle images into an Euclidean space, where the distance can directly measure the similarity of two vehicle images. Liu  \emph{et al.}\cite{liu2016deep} also constructed another dataset called VeRi-776, which adopts visual features, license plates and spatial-temporal information, to explore the vehicle Re-ID task. Shen \emph{et al.}\cite{shen2017learning} proposed a two-stage framework that incorporates complex spatial-temporal information of vehicles to effectively regularize Re-ID results. Yan \emph{et al.}\cite{yan2017exploiting} presented two high-quality vehicle datasets, namely, VD1 and VD2, which contain diverse annotated attributes. They also implemented generalized pairwise and multigrain-based list rankings with multi-attribute classification in a multitask deep learning framework to handle vehicle Re-ID task. Zhou \emph{et al.}\cite{zhou2018viewpoint} designed a viewpoint-aware attentive multi-view inference (VAMI) model that only requires visual information to solve multi-view vehicle Re-ID problems. While He \emph{et al.}\cite{he2019part} proposed a simple yet efficient part-regularized discriminative feature-preserving method, which enhances the perceptive capability of subtle discrepancies, and reported promising improvement.
\begin{figure*}[ht]
\centering 
\includegraphics[scale=0.53]{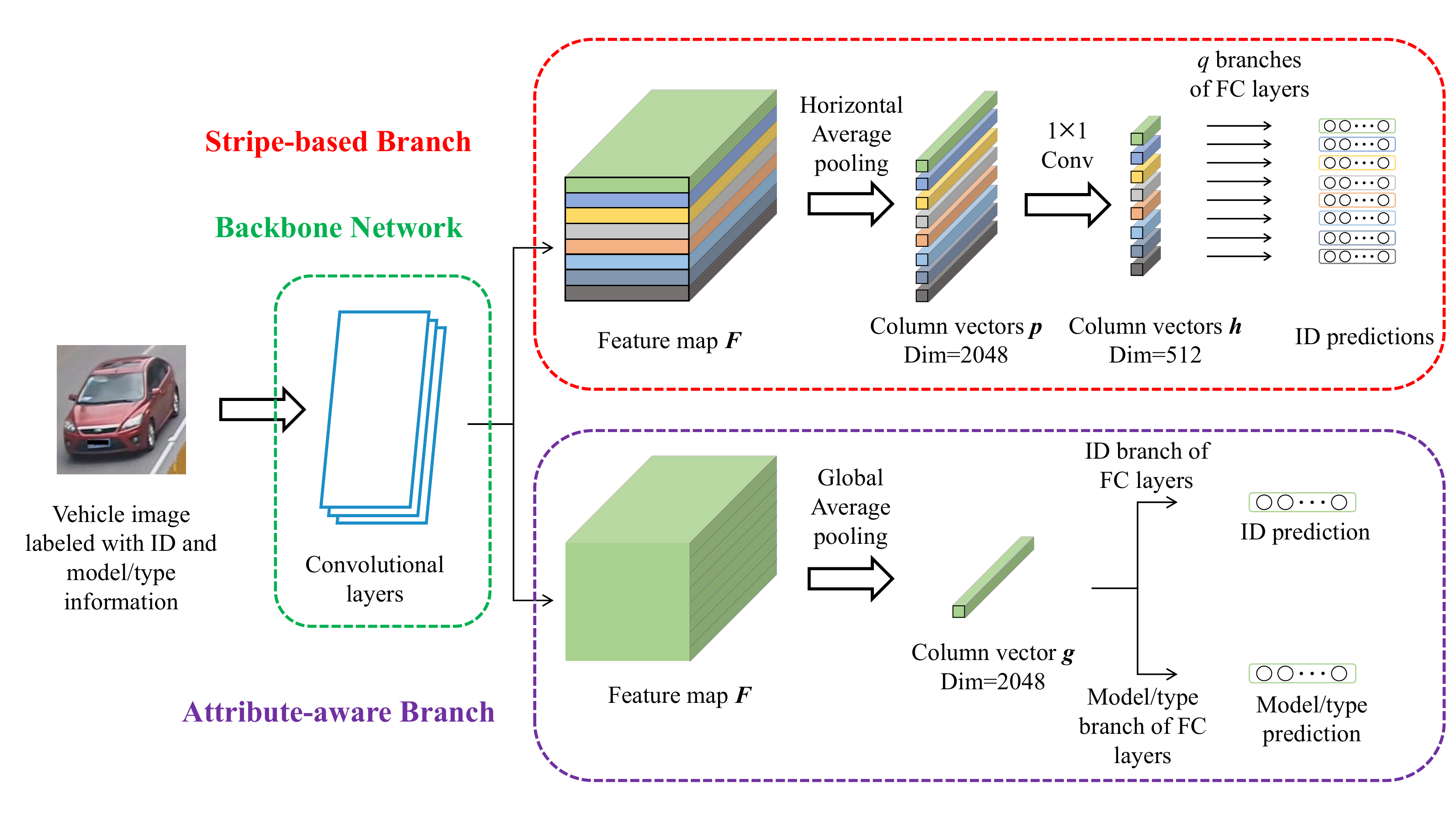} 
\caption{Diagram of the proposed SAN. In this paper, ResNet-50 is used as the backbone network, which is split into two independent branches that use the same feature map \textbf{\emph{F}} generated by stacked convolutional layers from the backbone network. In the stripe-based branch, $1 \times 1$ Conv refers to 1*1 convolutional layer for reducing the dimension of \textbf{\emph{p}}. Each dimension-reduced column vector \textbf{\emph{h}} is inputted into an ID classifier, whereas column vector \textbf{\emph{g}} is fed into an ID and model/type classifiers. Each classifier in stripe-based and attribute-aware branches is implemented with a fully-connected (FC) layer and a sequential Softmax layer. During training, each classifier is supervised by cross-entropy loss and predicts the identity or model/type of the input vehicle image. During testing, \emph{q} pieces of \textbf{\emph{h}} and \textbf{\emph{g}} are concatenated to form the final descriptor of the input vehicle image.} 
\label{Fig1}
\end{figure*}
\subsection{Discriminative part-level features}
Many recent works\cite{Wang2017Orientation, khorramshahi2019attention, khorramshahi2019dual, zhou2018viewpoint, he2019part, wang2019vehicle, guo2019two, teng2018scan, zhu2019vehicle} in vehicle Re-ID have used discriminative part-level features and reported encouraging improvement. After deep learning methods dominated the computer vision community, handcrafted part features for fine-grained recognition have declined. Therefore, extracting discriminative part-level features through deep learning methods is widely used by the computer vision community. Some works\cite{Wang2017Orientation, khorramshahi2019attention, khorramshahi2019dual} in vehicle Re-ID have utilized vehicle key points to learn local region features. Wang \emph{et al.}\cite{Wang2017Orientation} proposed to exploit 20 key point locations to localize and extract discriminative part-level features. Zhou \emph{et al.}\cite{zhou2018viewpoint} adopted a viewpoint-aware attention model to select the core regions at different viewpoints of vehicle images. They implemented an effective multi-view feature inference by extracting the discriminative part-level features of the selected core regions.
Several recent works\cite{he2019part, wang2019vehicle, guo2019two, teng2018scan} in vehicle Re-ID have stated that specific parts such as windscreen, lights and vehicle brand tend to have much discriminative information. Guo \emph{et al.}\cite{guo2019two} proposed a novel two-level attention network supervised by a multigrain ranking loss (TAMR) to learn an efficient feature from windscreen and car-head parts. In \cite{zhu2019vehicle}, different directional part features were utilized for spatial normalization and concatenation to serve as a directional deep learning feature for vehicle Re-ID.
\subsection{Attribute-embedded Re-ID}
Attributes have been extensively investigated as the mid-level semantic information to boost vehicle Re-ID task. Several works\cite{li2017deep, zhou2018viewpoint, hou2019multi} in vehicle Re-ID have adopted vehicle attributes, such as model and color, to recognize vehicles as important traits. Li \emph{et al.}\cite{li2017deep} introduced the attribute recognition into a vehicle Re-ID framework, along with verification and triplet losses. 
Zheng \emph{et al.}\cite{zheng2019attributes} proposed a novel deep network architecture guided by meaningful attributes, including camera views, vehicle types, and colors, for vehicle Re-ID. Zhao \emph{et al.}\cite{zhao2019structural} collected a new vehicle dataset with 21 classes of structural attributes and proposed a novel region of interest (ROIs)-based vehicle Re-ID and retrieval method, in which the deep features of ROIs were extracted on the basis of structural attributes. In addition, re-ranking with attribute information has been used by several scholars\cite{nguyen2019vehicle, huang2019multi}. Nguyen \emph{et al.}\cite{nguyen2019vehicle} applied specialized attribute classifiers to narrow down vehicle retrieval results by focusing on specific attributes of a vehicle. Huang \emph{et al.}\cite{huang2019multi} proposed a viewpoint-aware temporal attention model for vehicle Re-ID, which utilizes deep learning features and metadata attributes from the re-ranking phase for result refinement.
\section{Proposed Method}
\subsection{SAN}
In this paper, we propose a two-branch deep convolutional neural network to learn efficient feature embedding for vehicle Re-ID. The overall framework of the proposed method is shown in Fig. \ref {Fig1}. Given as input a vehicle image, the stacked convolutional layers from the backbone network will automatically form a 3D feature map. A conventional pooling layer is inserted into the stripe-based branch instead of the original global pooling layer, to spatially down-sample the feature map into pieces of column vectors. Then, the column vectors will be dimension-reduced and used to achieve part-level features. Meanwhile the attribute-aware branch extracts the global feature from the feature map under the supervision of vehicle attribute labels. The details of these components are described below.
\subsubsection{Backbone Network}
Diverse backbone networks (e.g., VGG\_{}CNN\_{}M\_{}1024 \cite{chatfield2014return}, Google Inception \cite{szegedy2017inception}, and ResNet \cite{he2016deep}) have been adopted in previous vehicle Re-ID models. In concept, SAN can take any deep convolutional neural network designed for image classification as the backbone network. In this paper, ResNet-50 is used due to its competitive performance and relatively concise architecture. Following the modification strategies in several state-of-the-art person Re-ID models \cite{sun2018beyond, luo2019bag}, the last spatial down-sampling operation in the backbone network is set to 1 to obtain a feature map with high spatial size. High spatial resolution always enriches the granularity of feature.
\begin{figure}[t]
\centering
\includegraphics[scale=0.35]{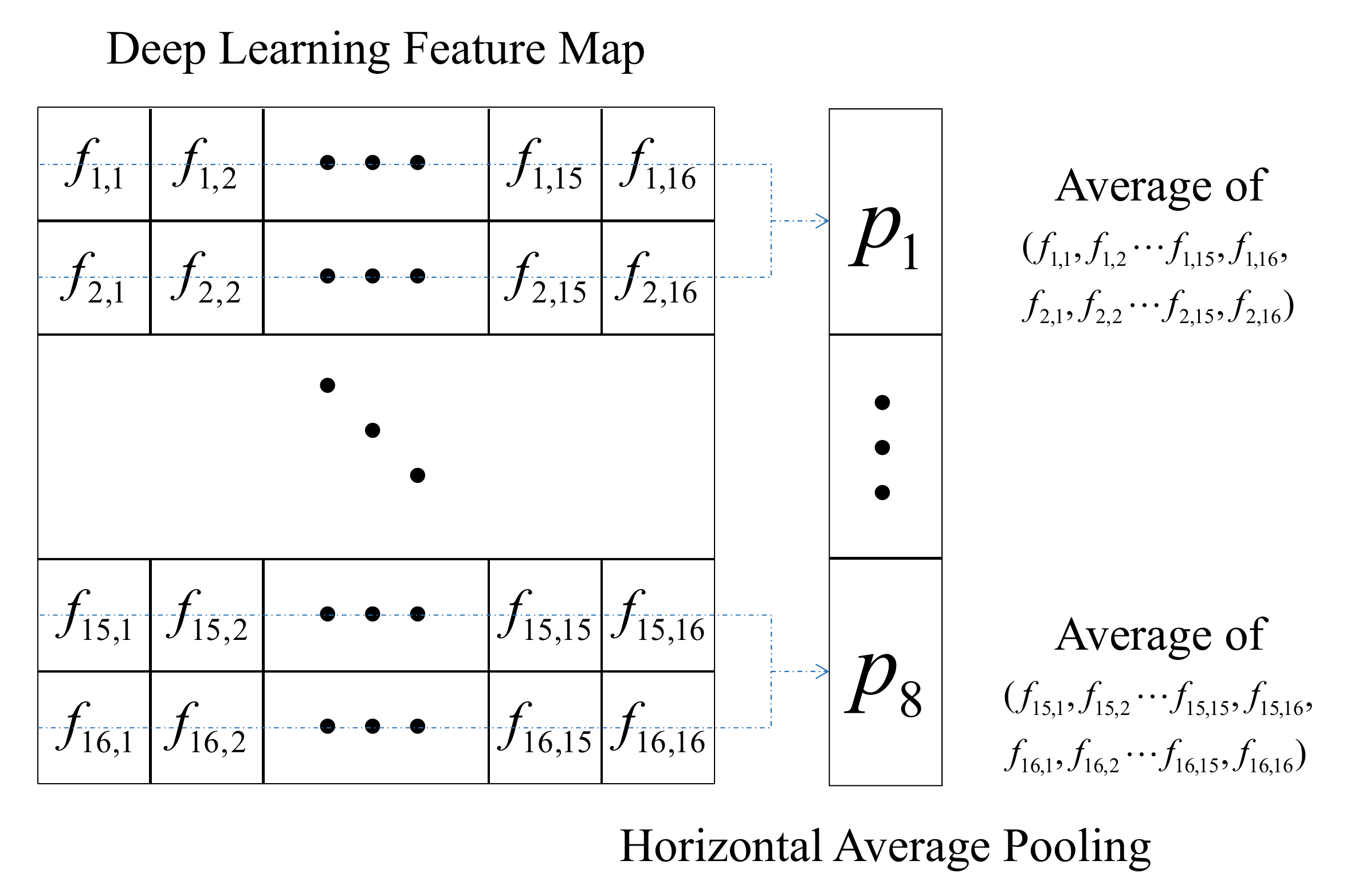}
\caption{Schematic diagram of horizontal average pooling operation.}
\label{Fig2}
\end{figure}
\subsubsection{Stripe-based Branch}
Height-wise partition is a common strategy in person Re-ID models, because the human body can be divided into several meaningful parts (e.g., head, thorax, legs, and feet). Similar to a human body, a car body can be roughly divided into several meaningful parts in the vertical axis (e.g., ceiling, windshield, header panel, and wheels). Therefore, a horizontal average pooling layer and a dimension-reduced convolutional layer in the stripe-based branch are designed to extract the part-level features of vehicle images in the vertical axis. The basic feature map of activations produced by the backbone network is assumed to be $\bm{F}\in \mathbb{R}^{m\times m\times c}$, where $m$ and $c$ represent the height/width, and channel sizes, respectively. In addition, the vector of activations viewed along the channel axis is defined as a \textbf{column vector}. The horizontal average pooling and dimension-reduced convolutional layers can be described as follows.
\begin{enumerate}[label=\roman*), listparindent=0.3em]
\item
\textbf{Horizontal Average Pooling Layer}: The horizontal average pooling layer partitions $\bm{F}$ into $q$ horizontal stripes and averages all column vectors in a same stripe into a single part-level column vector $\bm{p}_i$ ($i = 1,2,\cdots,q$, the subscripts will be omitted unless necessary). Combine all part-level column vectors together to obtain the horizontal average pooling feature map $\bm{P}\in \mathbb{R}^{q\times 1\times c}$, as shown in Fig. \ref {Fig2}. For example, $p_1$ is equal to the average of $f_{1,1},f_{1,2} \cdots f_{1,15},f_{1,16},f_{2,1},f_{2,2} \cdots f_{2,15},f_{2,16}$, that is, $p_1 = \frac{1}{32}(f_{1,1}+f_{1,2}+\cdots  +f_{1,15}+f_{1,16}+f_{2,1}+f_{2,2}+\cdots +f_{2,15}+f_{2,16})$.
\item
\textbf{Dimension-reduced Convolutional Layer}: If each part-level column vector $\bm{p}$ is directly inputted to a classifier, the vehicle Re-ID model will incur too high computational resources. Moreover, several state-of-the-art person Re-ID works \cite{sun2018beyond, chen2019partition} have attempted to insert a convolutional layer before the classifier to reduce feature dimension and reported promising results. On the basis of such modification strategies, a dimension-reduced convolutional layer is used in the proposed network to reduce the dimension of $\bm{p}$, and the dimension-reduced column vectors $\bm{h}$ are set to 512-dim.
\end{enumerate}
\par
Given that the horizontal average pooling layer uses an average pooling operation, the forward and backward propagations can be briefly introduced as follows. Assume that the input feature map of an average pooling layer is $F = [F_1,F_2, \cdots ,F_m]^T \in \mathbb{R}^{m\times m}$, the average pooling window size is $2 \times m$, and the output feature map is $P \in  \mathbb{R}^{ \frac{m}{2}\times 1} = [p_1,p_2, \cdots ,p_{\frac{m}{2}}]^T$. Then, the forward propagation of this average pooling layer can be calculated as follows:
\begin{equation}
p_i = \frac{1}{2m}\sum_{j=2i-1}^{2i}\sum_{k=1}^mF_{j,k},
\end{equation}
where $p_i$ is the $i$-th element of $P$, and $F_{j,k}$ is the $k$-th element of the $j$-th row vector of $F$. On the basis of the chain rule, the backward propagation of this average pooling layer can be calculated as follows:
\begin{gather}
\frac{\partial L}{\partial F_{2i-1,1}} = \frac{\partial L}{\partial p_i} \frac{\partial p_i}{\partial F_{2i-1,1}} = \frac{1}{2m}\frac{\partial L}{\partial p_i}, \nonumber \\
\dots \nonumber \\
\frac{\partial L}{\partial F_{2i-1,m}} = \frac{\partial L}{\partial p_i} \frac{\partial p_i}{\partial F_{2i-1,m}} = \frac{1}{2m}\frac{\partial L}{\partial p_i}, \nonumber \\
\frac{\partial L}{\partial F_{2i,1}} = \frac{\partial L}{\partial p_i} \frac{\partial p_i}{\partial F_{2i,1}} = \frac{1}{2m}\frac{\partial L}{\partial p_i}, \nonumber \\
\dots \nonumber \\
\frac{\partial L}{\partial F_{2i,m}} = \frac{\partial L}{\partial p_i} \frac{\partial p_i}{\partial F_{2i,m}} = \frac{1}{2m}\frac{\partial L}{\partial p_i},
\end{gather}
and the matrix form can be formulated as follows:
\begin{gather}
\frac{\partial L}{\partial F_{2i-1}} = \frac{1}{2m}[\frac{\partial L}{\partial p_i},\frac{\partial L}{\partial p_i},\cdots,\frac{\partial L}{\partial p_i}], \nonumber \\
\frac{\partial L}{\partial F_{2i}} = \frac{1}{2m}[\frac{\partial L}{\partial p_i},\frac{\partial L}{\partial p_i},\cdots,\frac{\partial L}{\partial p_i}],
\end{gather}
where $F_{2i-1} = [F_{2i-1,1},F_{2i-1,2},\cdots,F_{2i-1,m}]$ is the ($2i-1$)-th row vector of $F$, and $L$ is the loss function (i.e., Eq. \eqref{loss}) of the overall learning framework, which will be discussed in the following subsection.
\subsubsection{Attribute-aware Branch}
Given that the vehicle's attributes (i.e., ID, model, and type) are not absolutely mutually exclusive, multi-attribute prediction essentially becomes a multi-label classification problem. In the attribute-aware branch, a global average pooling layer is used to obtain the global features of the input vehicle images. After the global average pooling layer, the branch is divided into two parts. Each part contains a fully-connected layer, with $K$ neuron units to predict $K$-classes, respectively.
\par
The vehicle's attributes are utilized in the attribute-aware branch to extract the global feature. However, not all the images in the VehicleID dataset are labeled with the vehicle model attribute; only 90186 vehicle images in the training set are annotated with model information while the others are not. To solve the missing model label problem, DenseNet121 \cite{huang2017densely} is adopted to predict the model labels of the vehicle images without model annotations. During the training phase, the DenseNet121 model is trained by images with ID and model annotations and supervised by cross-entropy loss to predict the identity and model of the input vehicle image. During the testing phase, the DenseNet121 model predicts the \textbf{soft-label} of the vehicle image without model annotation. Here, the model label of vehicle image is predicted by the model instead of being manually labeled, hence, we name the predicted model label of the vehicle image as soft-label.
\par
On the basis of the aforementioned backbone network, stripe-based branch and attribute-aware branch, the SAN is constructed, which consists of horizontal average pooling and dimension-reduced convolutional Layers, as shown in Fig. \ref {Fig1}.
\par
Last but not least, the effectiveness of the horizontal average pooling layer in suppressing the horizontal viewpoint variations can be explained as follows. In the proposed method, the backbone network transforms an input vehicle image $I$ with size $256\times 256\times 3$ (i.e., height $\times$ width $\times$ channel) into a feature map $F$ of $16\times 16\times 2048$. For the feature map $F$, the horizontal average pooling layer further obtains the average of every two rows of $F$ to form the output feature map $P$ of  $8\times 1\times 2048$. 
Each element of the output feature map $P$ is a stable feature (i.e., the mean value of every two rows) calculated from a large horizontal stripe reception field (i.e., height $\times$ width = $32 \times 256$), which covers the entire input image in the horizontal direction. Consequently, the proposed horizontal average pooling layer becomes more robust to the appearance changes caused by horizontal viewpoint variations. Therefore, the proposed method can comprehensively describe an input image in the horizontal direction by using a horizontal average pooling layer.
\subsection{Loss Function}
Similar to \cite{hou2019multi} and \cite{chen2019partition}, the Softmax function is utilized to predict the identity or model/type of the input vehicle image. Assuming that the input vehicle image is $I$, the vehicle image with model annotation is $I_m$, and the vehicle image with model soft-label is $I_s$. Here, the loss function of the proposed method can be expressed as follows:
\begin{equation}
L = \sum_{i=1}^{q}L_i + L_{ID} + L_{Model},
\label{loss}
\end{equation}
where $q$ is the number of stripes; $L_i$ represents the cross-entropy loss of the $i$-th stripe in the stripe-based branch; and $L_{ID}$ and $L_{Model}$ are the cross-entropy losses of the ID and model/type branches in the attribute-aware branch, respectively. In our experiments, $L_i$, $L_{ID}$ and $L_{Model}$ are formulated as follows:
\begin{equation}
L_i = -\sum_{k=1}^{K_i}log(p(k_i))q(k_i),
\label{l1}
\end{equation}
\begin{equation}
L_{ID} = -\sum_{c=1}^{C}log(p(c))q(c),
\label{l2}
\end{equation}
\begin{equation}
L_{Model} = \left\{  
             \begin{array}{clr}  
             -\sum_{m=1}^{M}log(p(m))q(m), I = I_m \\
             -\sum_{m=1}^{M}log(p(m))s(m), I = I_s,
             \end{array}  
\right.
\label{l3}
\end{equation}
where $K_i$ is the class number of $i$-th stripe; $C$ and $M$ are the class numbers of the ID and model/type branches in the attribute-aware branch, respectively; $p(k_i), p(c), p(m) \in [0,1]$ are the $i$-th stripe prediction probability of class $k$, the ID branch prediction probability of class $c$, and the model/type branch prediction probability of class $m$, respectively; $q(k_i), q(c)$ are the ground truth label distribution of the $i$-th stripe and ID branch; and $q(m), s(m)$ are the ground truth label distribution and the ground truth soft-label distribution of the model/type branch, respectively. $q(k_i), q(m), s(m)$ can be formulated as follows ($q(c)$ can be calculated by Eq. \eqref{q1} as well):
\begin{equation}  
q(k_i)= \left\{  
             \begin{array}{clr}  
             0,    k_i \neq y_i &  \\  
             1,    k_i = y_i, &    
             \end{array}  
\right.
\label{q1}
\end{equation} 
\begin{equation}  
q(m)= \left\{  
             \begin{array}{clr}  
             0,    m \neq y_m &  \\  
             1,    m = y_m, &    
             \end{array}  
\right.
\end{equation} 
\begin{equation}  
s(m)= \left\{  
             \begin{array}{clr}  
             0,    m \neq y_s &  \\  
             1,    m = y_s, &    
             \end{array}  
\right.
\end{equation} 
where $y_i$ is the ground truth ID label, $y_m$ is the ground truth model label, and $y_s$ is the ground truth model soft-label. All terms, except the ground truth term, will not be considered in Eq. \eqref{l1}\eqref{l2}\eqref{l3}. Thus, cross-entropy losses $L_i$, $L_{ID}$, and $L_{Model}$ can be simplified as:
\begin{equation}
L_i = -log(p(y_i)),
\end{equation}
\begin{equation}
L_{ID} = -log(p(y_i)),
\end{equation}
\begin{equation}
L_{Model} = \left\{  
             \begin{array}{clr}  
             -log(p(y_m)), I = I_m  &\\
             -log(p(y_s)), I = I_s. &
             \end{array}  
\right.
\end{equation}
\section{Experiments}
In this section, we initially perform ablation studies on VehicleID \cite{liu2016deepdeep} and VeRi \cite{liu2016deep} datasets to validate the benefit of each branch of SAN both quantitatively and qualitatively. Then, we list the comparisons with various state-of-the-art vehicle Re-ID methods to demonstrate the superiority of SAN. In our experiments, the Euclidean distance is used to measure the similarity of a vehicle pair described using part-level and global features.
\makeatletter
\def\hlinew#1{
  \noalign{\ifnum0=`}\fi\hrule \@height #1 \futurelet
   \reserved@a\@xhline}
\makeatother

\begin{table*}[ht]
\renewcommand{\arraystretch}{1.3}
\renewcommand{\multirowsetup}{\centering}
\caption{Detailed results (\%) of ablation study on VehicleID and VeRi. Components marked with ``\checkmark" are used in each model. Here, $L_{ID}$ stands for the attribute-aware branch only with ID cross-entropy loss, whereas $L_{Model}$ stands for the attribute-aware branch only with model cross-entropy loss.}
\label{ablation table}
\centering
\begin{tabular}{c|ccc|cc|cc|cc|cc|ccc}
\hlinew{1.5pt}
\multirow{3}{*}{Settings}  & \multicolumn{3}{c|}{Striped-based Branch}                                                                                         & \multicolumn{2}{c|}{attribute-aware Branch}                       & \multicolumn{6}{c|}{VehicleID}                                                                  & \multicolumn{3}{c}{VeRi}                                          \\ \cline{2-15} 
                           & \multirow{2}{0.7cm}{q=2} & \multirow{2}{0.7cm}{q=4} & \multicolumn{1}{c|}{\multirow{2}{0.7cm}{q=8}} & \multirow{2}{1.1cm}{$L_{ID}$} & \multirow{2}{0.9cm}{$L_{Model}$} & \multicolumn{2}{c|}{Test = 800} & \multicolumn{2}{c|}{Test = 1600} & \multicolumn{2}{c|}{Test = 2400} & \multirow{2}{*}{mAP} & \multirow{2}{*}{r=1} & \multirow{2}{*}{r=5} \\
                           &                      &                      & \multicolumn{1}{c|}{}                     &                     &                        & r=1           & r=5           & r=1            & r=5           & r=1            & r=5           &                      &                      &                      \\ \hlinew{1pt}
\multirow{6}{*}{ResNet-50} &                                           &                                           &                                           & \checkmark                                        &                        & 68.0          & 82.1          & 66.9           & 79.6          & 62.2           & 76.8          & 61.0                 & 89.2                 & 95.5                 \\
&                                           &                                           &                                           & \checkmark                                        & \checkmark                      & 76.1          & 90.9          & 73.7           & 88.0          & 70.9           & 85.4          & 66.7                 & 90.3                 & 95.8                 \\
                           & \checkmark                                         &                                           &                                           &                                          &                        & 74.0          & 88.2          & 73.3           & 85.7          & 71.6           & 82.7          & 67.0                 & 92.1                 & 96.4                 \\
                           &                                           & \checkmark                                         &                                           &                                          &                        & 74.6          & 90.2          & 74.1           & 87.0          & 72.6           & 83.2          & 67.4                 & 92.1                 & 96.4                 \\
                           &                                           &                                           & \checkmark                                         &                                          &                        & 76.1          & 91.6          & 75.1           & 87.4          & 73.4           & 84.9          & 70.1                 & 93.2                 & 97.0                 \\
                           &                                           &                                           & \checkmark                                         & \checkmark                                        &  \checkmark                      & \textbf{79.7}          & \textbf{94.3}          & \textbf{78.4}           & \textbf{91.3}          & \textbf{75.6}           & \textbf{88.3}          & \textbf{72.5}                 & \textbf{93.3}                 & \textbf{97.1}                 \\ \hlinew{1.5pt}
\end{tabular}
\end{table*}

\begin{figure*}[ht]
\centering
\subfigure[]{
\begin{minipage}[t]{0.24\linewidth}
\centering
\includegraphics[scale=0.24]{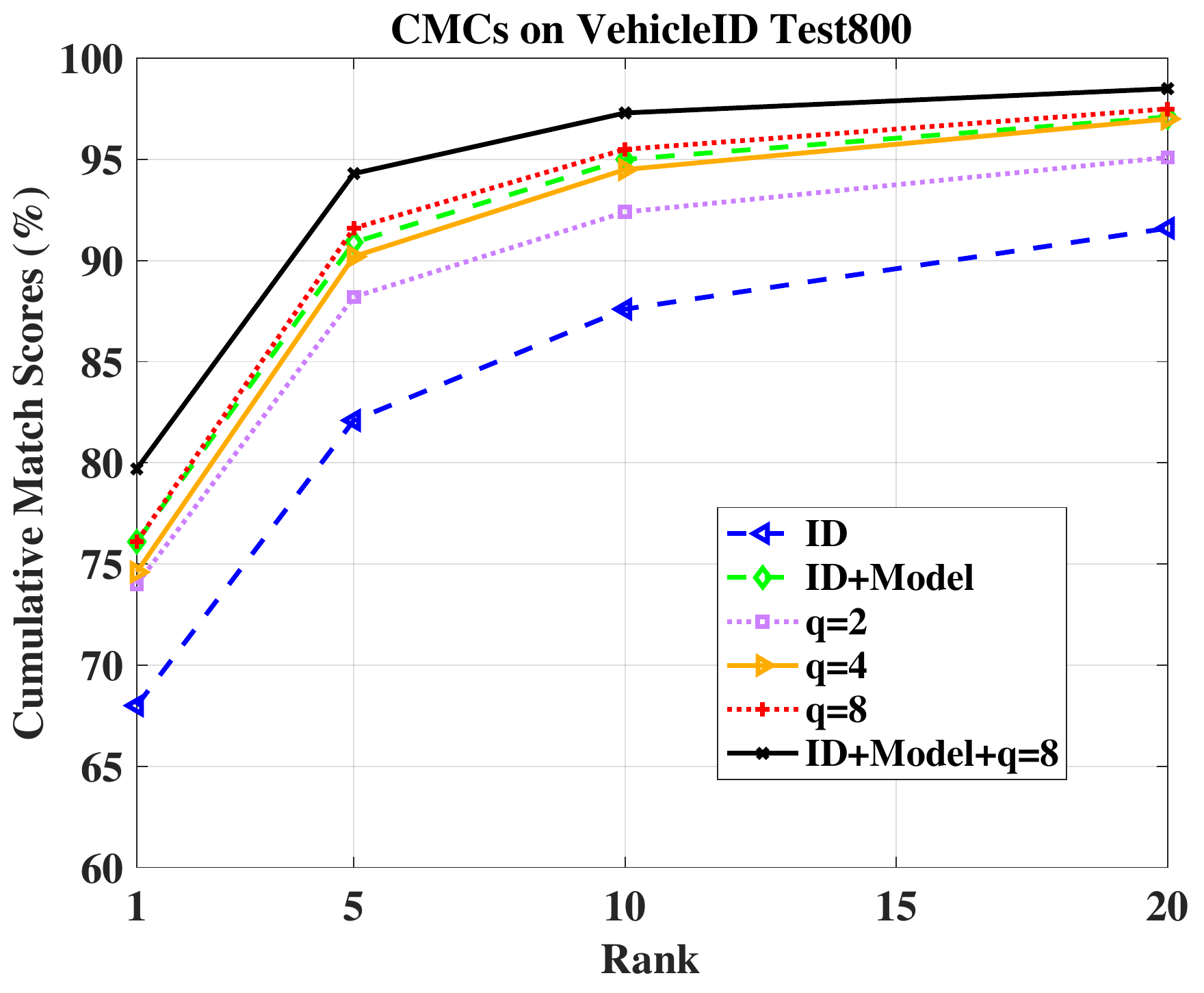}
\end{minipage}%
}%
\subfigure[]{
\begin{minipage}[t]{0.24\linewidth}
\centering
\includegraphics[scale=0.24]{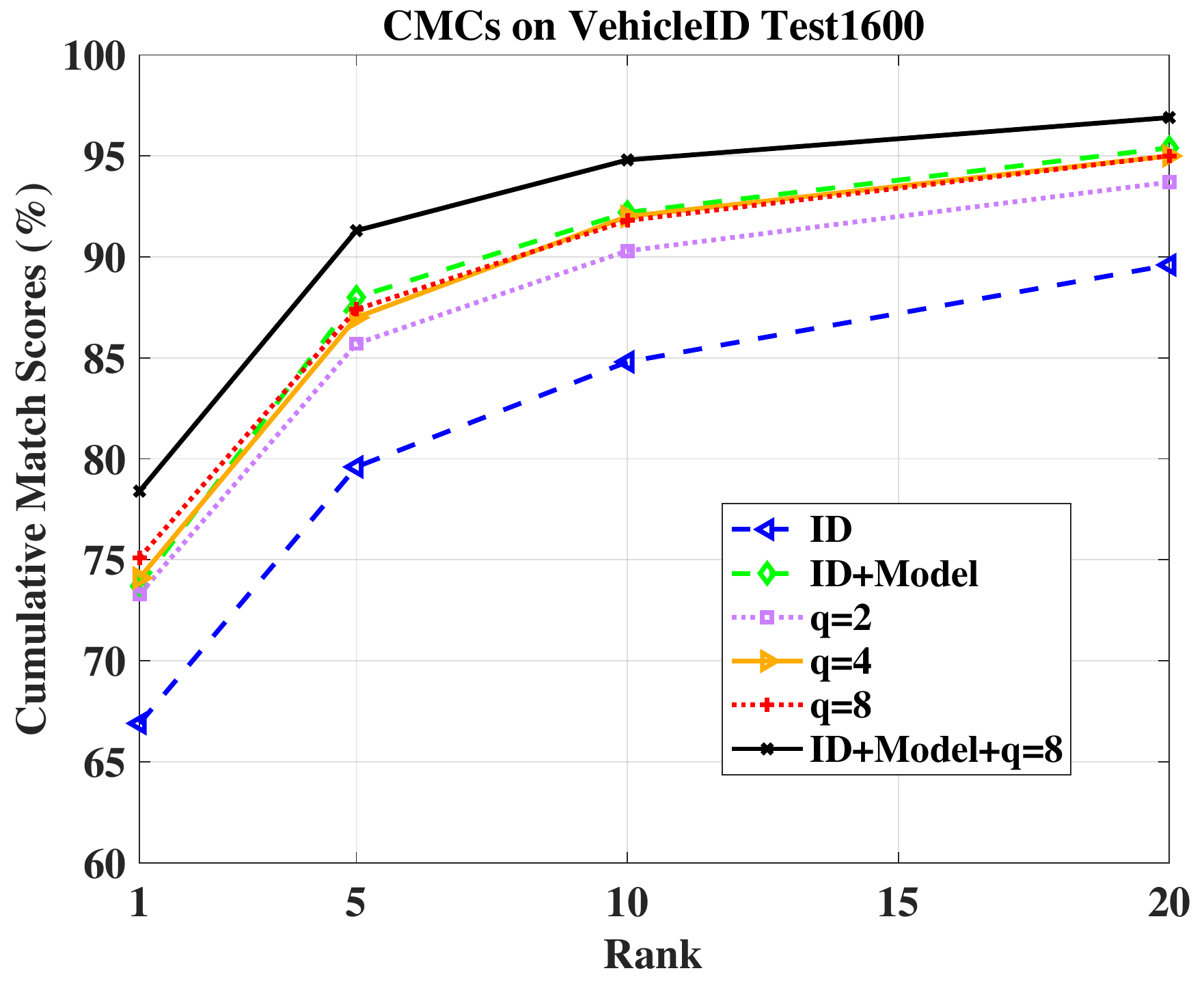}
\end{minipage}%
}%
\subfigure[]{
\begin{minipage}[t]{0.24\linewidth}
\centering
\includegraphics[scale=0.24]{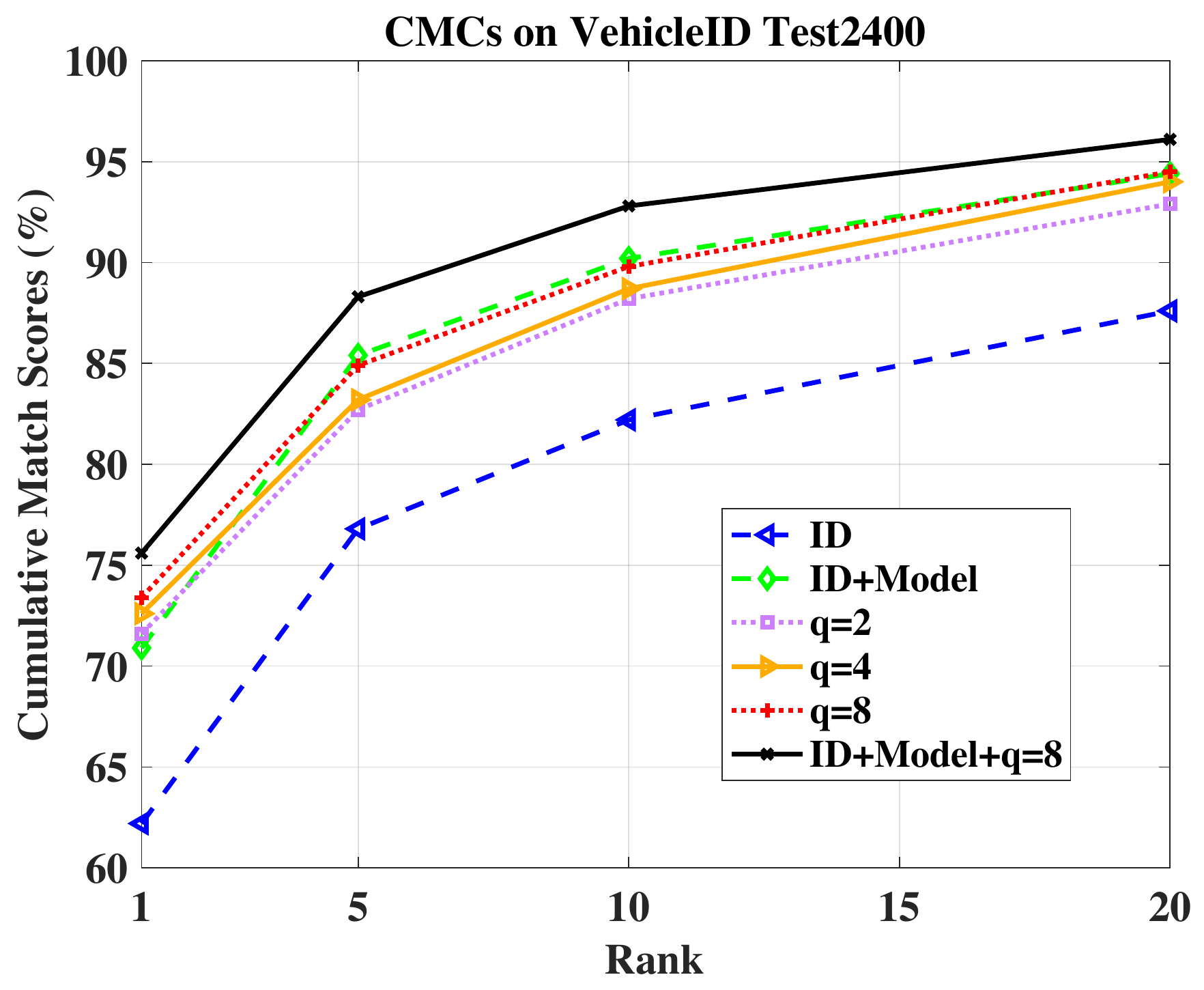}
\end{minipage}
}%
\subfigure[]{
\begin{minipage}[t]{0.24\linewidth}
\centering
\includegraphics[scale=0.24]{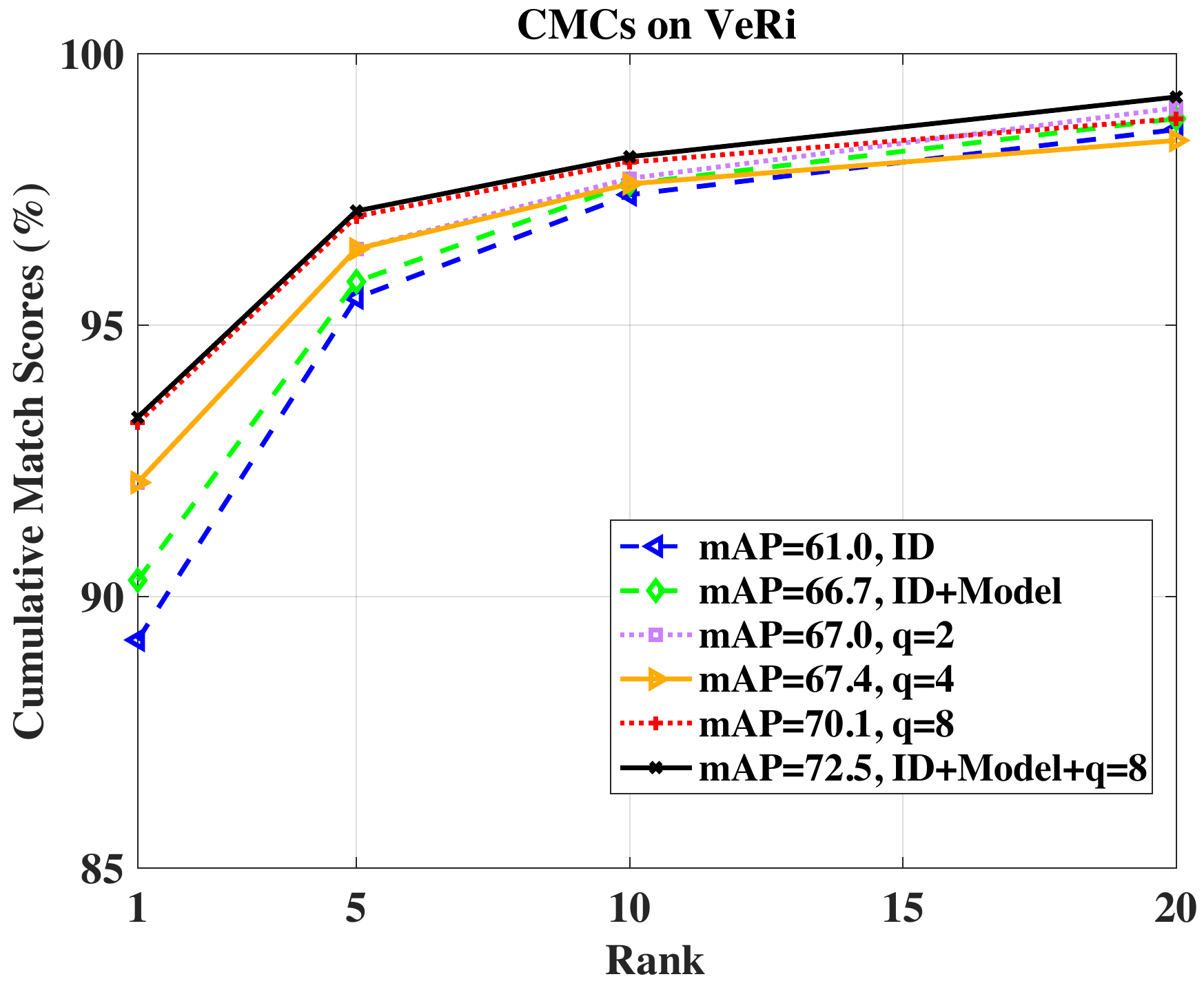}
\end{minipage}
}%
\caption{Cumulative matching characteristic (CMC) curve of ablation study on (a) Test800, (b) Test1600, and (c) Test2400 of VehicleID and (d) VeRi.}
\label{ablation figure}
\end{figure*}
\subsection{Datasets and Evaluation Protocols}
\textbf{Datasets.} We choose two of the most common vehicle datasets in our experiments, namely, VehicleID and VeRi:
\par
\emph{VehicleID}: This dataset contains 221,763 daytime images of 26,267 vehicles captured by multiple real-world surveillance cameras distributed in a small city of China. Each vehicle is captured from either the front or back viewpoint. All images are labelled with vehicle ID; however, not all the images are labelled with the vehicle model. A total of 250 models are annotated as attributes. The VehicleID dataset is divided into a training subset containing 110,178 images of 13,134 subjects and three testing subsets, namely, Test800, Test1600, and Test2400, to evaluate the performance in different data scales. Test800 contains 800 gallery and 6,532 probe images of 800 subjects. Test1600 comprises 1,600 gallery and 11,395 probe images of 1,600 subjects. Test2400 includes 2,400 gallery and 17,638 probe images of 2,400 subjects.
\par
\emph{VeRi}: This dataset contains over 50,000 images of 776 vehicles captured by 20 cameras in unconstrained traffic scenarios. Each vehicle is captured by 2-18 cameras under different viewpoints, illuminations, occlusions, and resolutions. The images are annotated with vehicle ID and type. Nine types are annotated as attributes. The VeRi dataset is divided into a training subset containing 37,781 images of 576 subjects and a testing subset with 13,257 images of 200 subjects. For the evaluation, one image of each vehicle captured from each camera is applied as a query. Then, a query set containing 1,678 images of 200 subjects and a gallery including 11,579 images of 200 subjects are finally obtained.
\par
\textbf{Evaluation Protocols.} Following the general evaluation protocols in the Re-ID field, the Cumulative Matching Characteristic (CMC) curve and mean Average Precision (mAP) are adopted to obtain a fair comparison with the existing methods. CMC is an estimation of finding the correct match in the top \emph{K} returned results. mAP is a comprehensive index which considers both the precision and recall of the results.
\begin{figure*}
\centering
\subfigure[Input image]{
\begin{minipage}[b]{0.17\linewidth}
\includegraphics[width=1\linewidth]{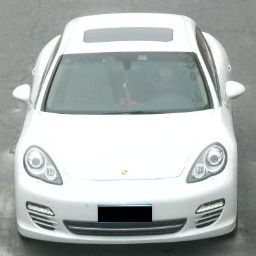}\vspace{4pt}
\includegraphics[width=1\linewidth]{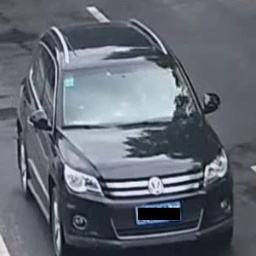}
\end{minipage}}
\subfigure[Baseline]{
\begin{minipage}[b]{0.17\linewidth}
\includegraphics[width=1\linewidth]{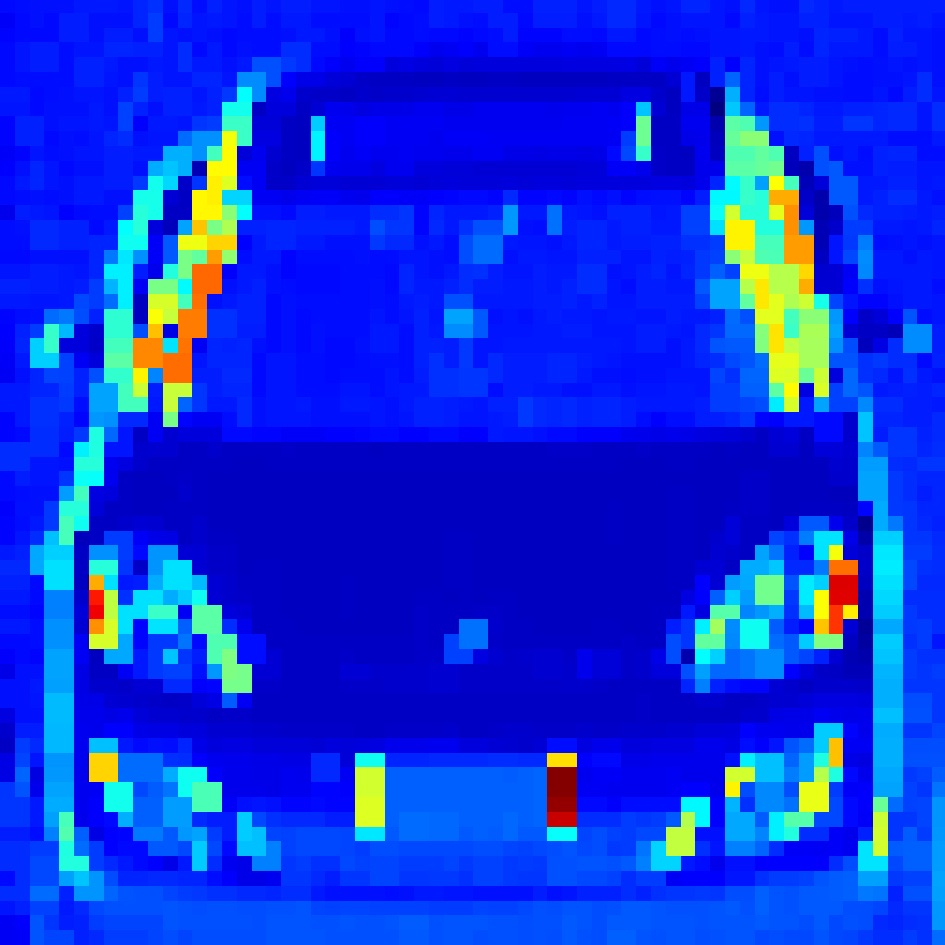}\vspace{4pt}
\includegraphics[width=1\linewidth]{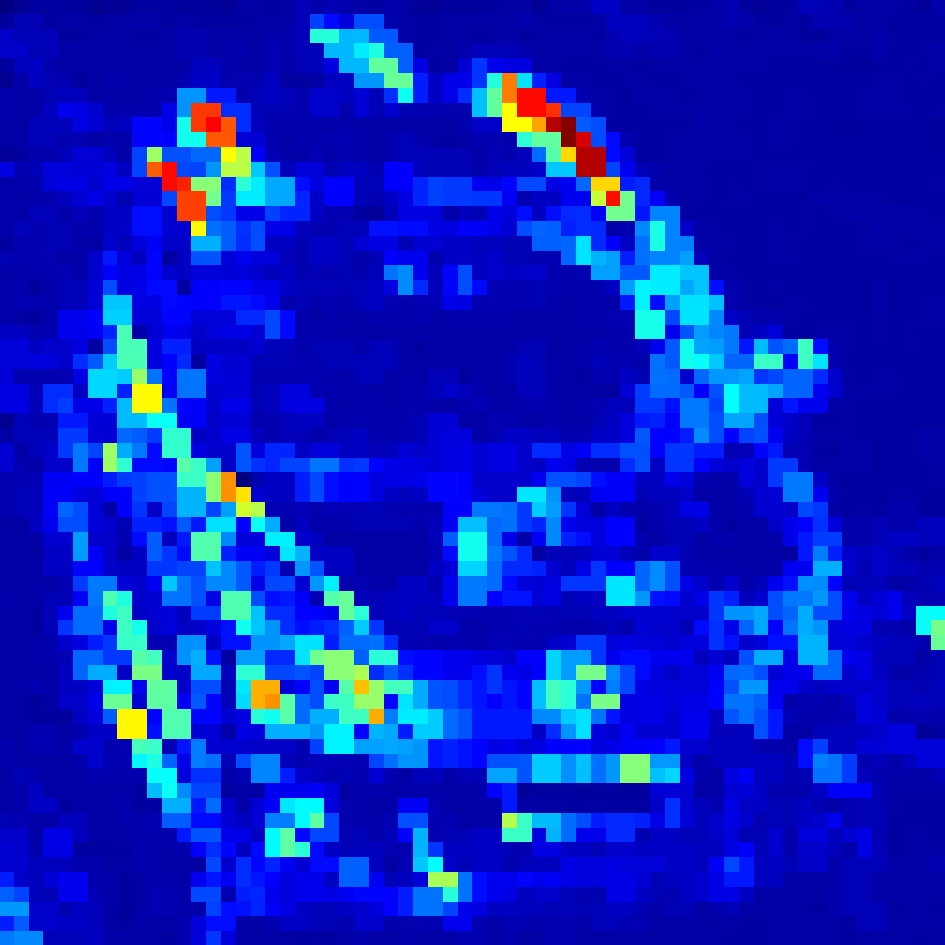}
\end{minipage}}
\subfigure[attribute-aware branch]{
\begin{minipage}[b]{0.17\linewidth}
\includegraphics[width=1\linewidth]{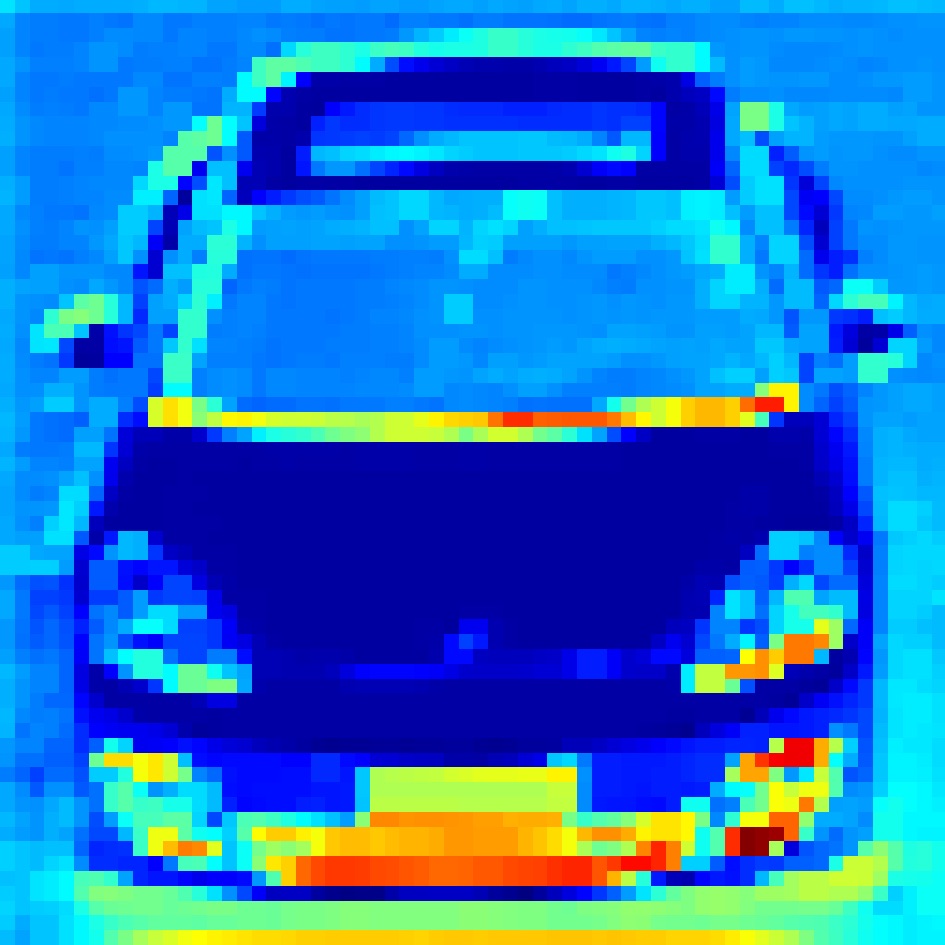}\vspace{4pt}
\includegraphics[width=1\linewidth]{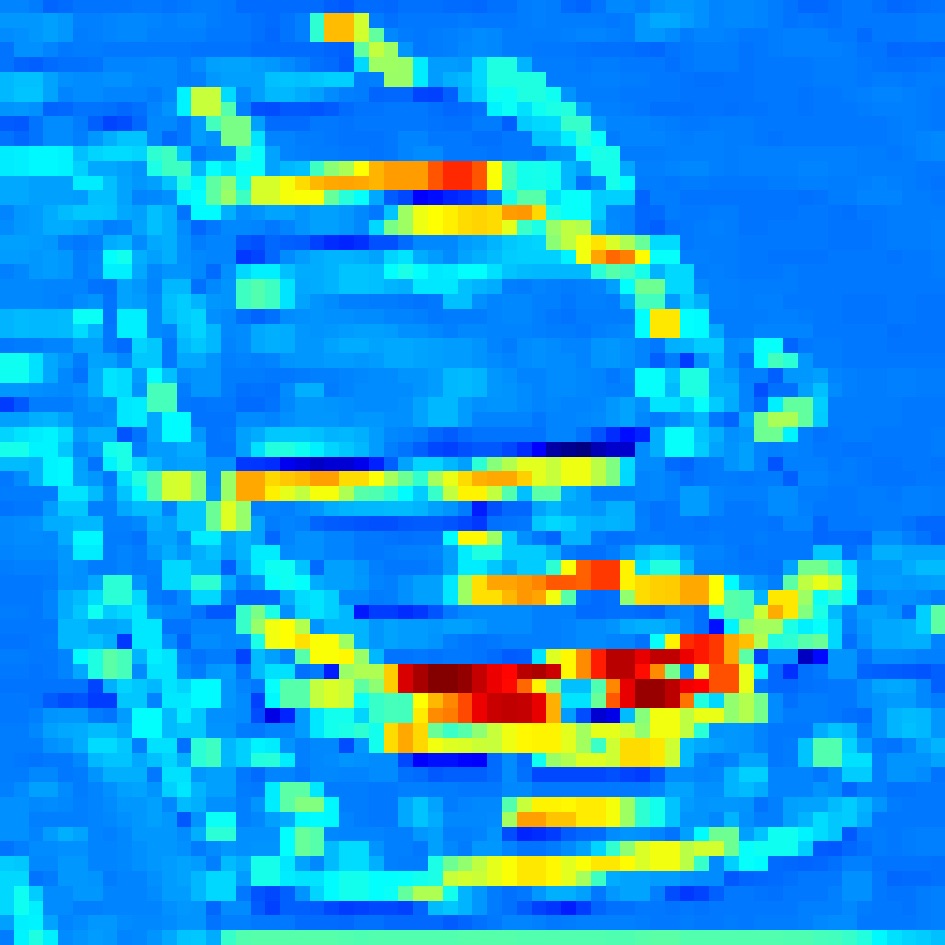}
\end{minipage}}
\subfigure[stripe-based branch]{
\begin{minipage}[b]{0.17\linewidth}
\includegraphics[width=1\linewidth]{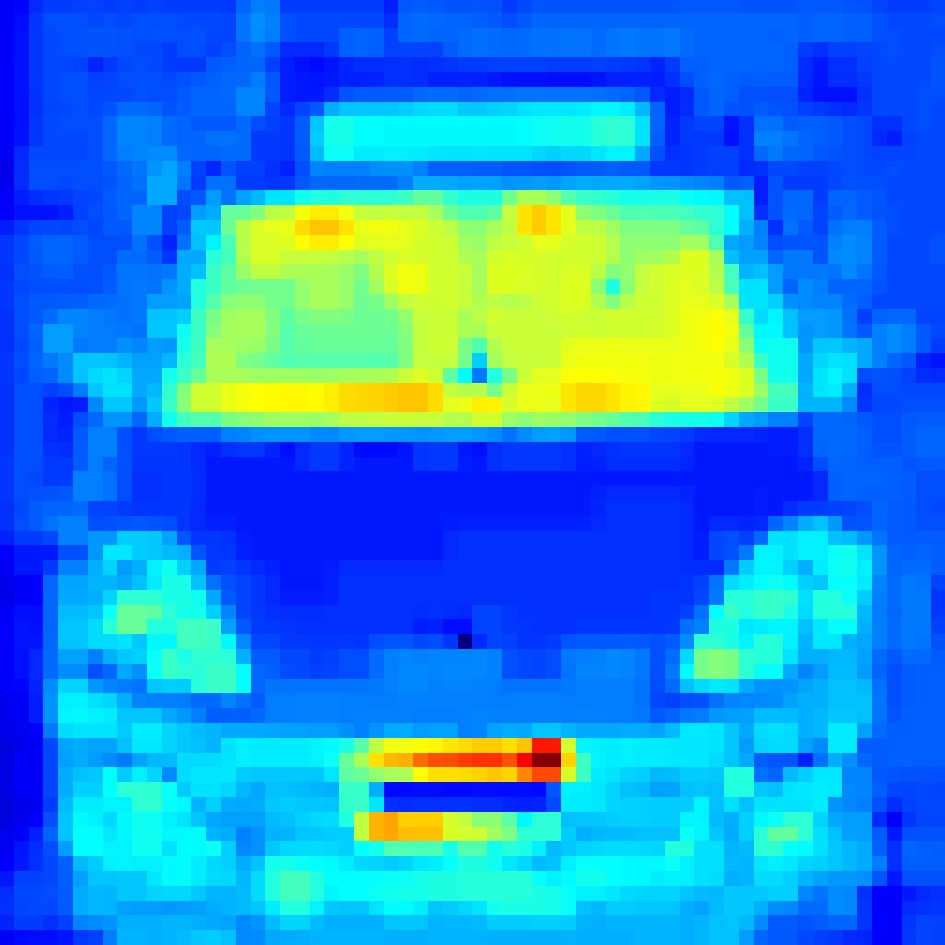}\vspace{4pt}
\includegraphics[width=1\linewidth]{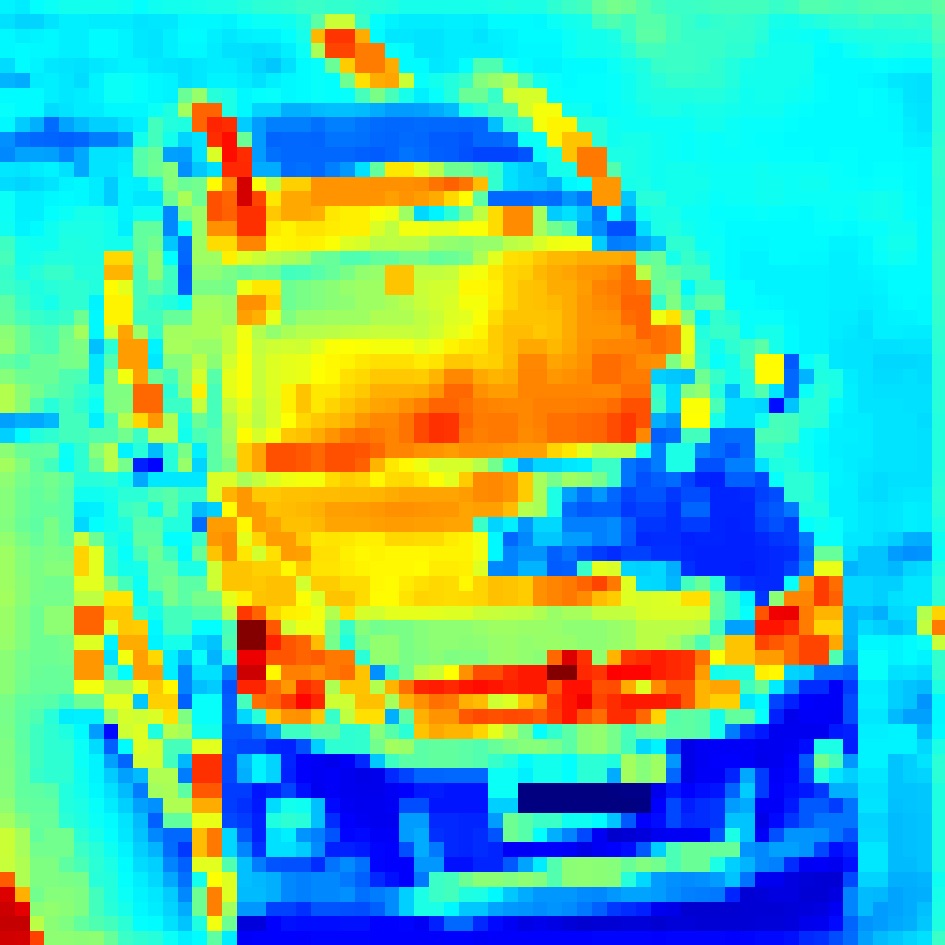}
\end{minipage}}
\subfigure[SAN]{
\begin{minipage}[b]{0.17\linewidth}
\includegraphics[width=1\linewidth]{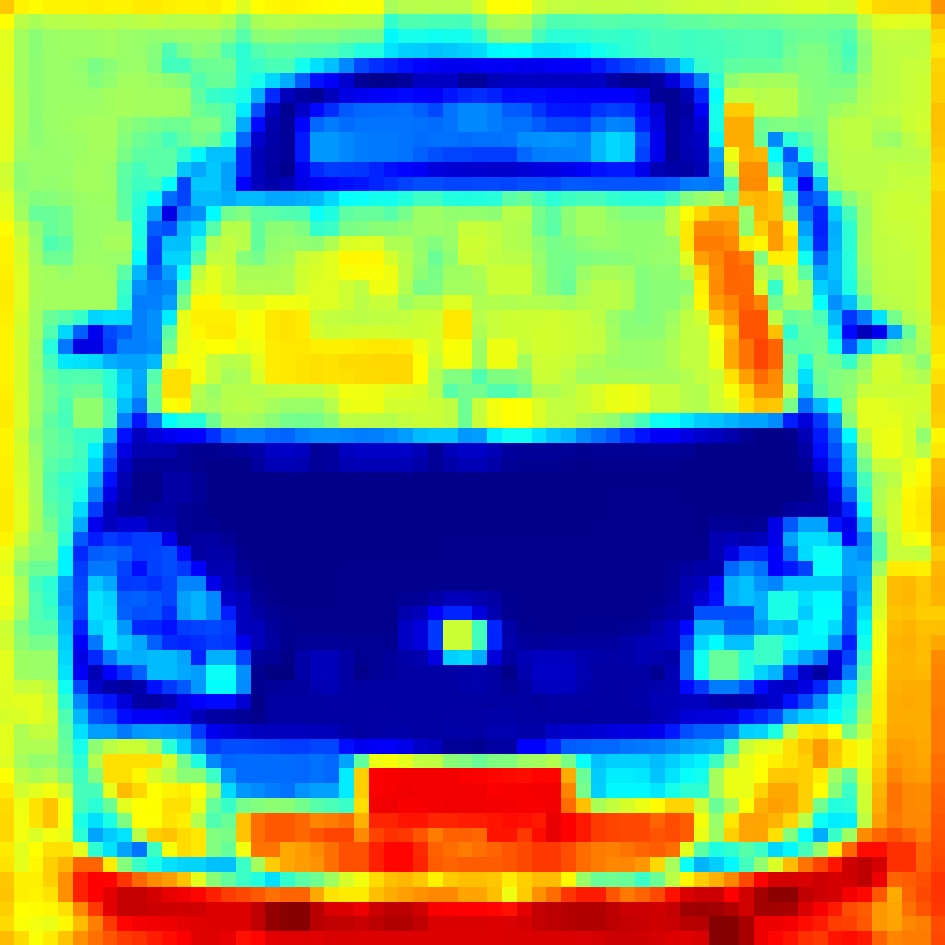}\vspace{4pt}
\includegraphics[width=1\linewidth]{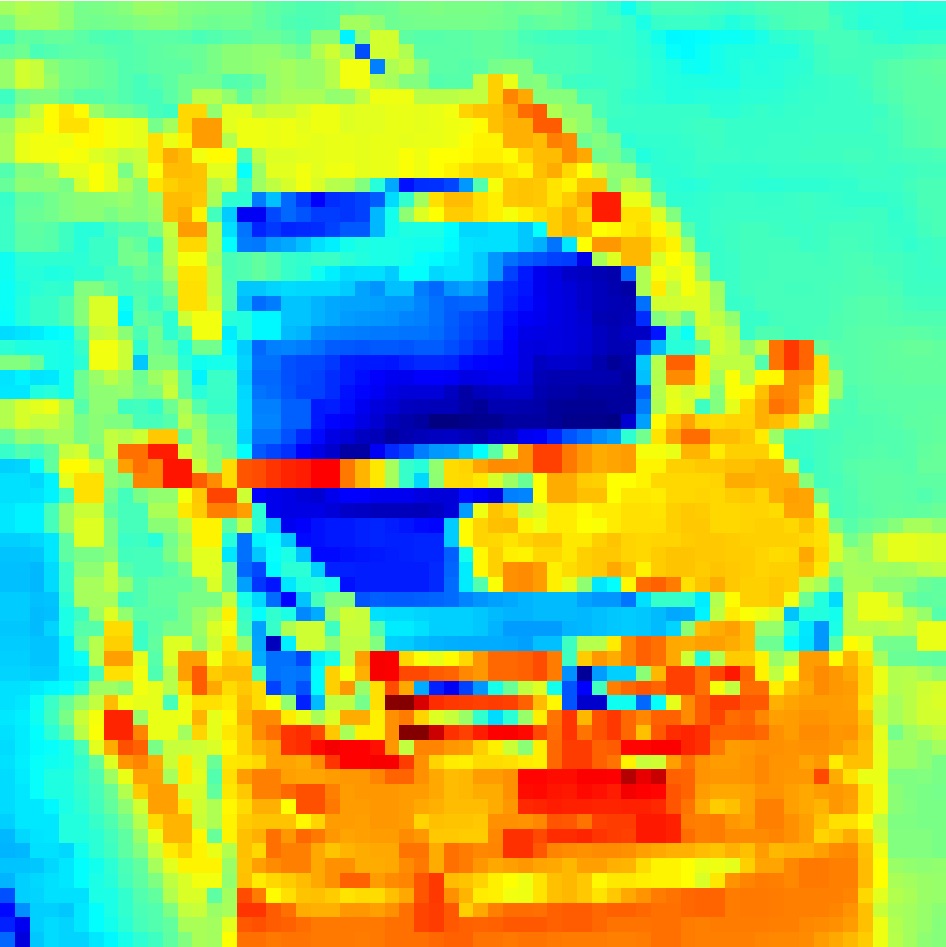}
\end{minipage}}
\caption{Visualization of the effectiveness of two-branch stripe-based and attribute-aware network. From left to right: input image, feature map of baseline only with $L_{ID}$, feature maps of attribute-aware and stripe-based branches, and feature map of the proposed SAN. In comparison with the baseline, the attribute-aware branch focuses more on the car-head part, which contains headlights, logo and license plate area. Local identity details, such as inspection marks on windscreen and use traces on car head, can be spotted only by the stripe-based branch. SAN, which consists of attribute-aware and stripe-based branches, further enhances these discriminative local details considerably.}
\label{visualization}
\end{figure*}
\subsection{Ablation Studies}
First, we conduct comparison experiments to validate the effectiveness of the attribute-aware branch. Then, we verify the benefits of the stripe-based branch. The detailed results of the ablation studies are illustrated in Table \ref {ablation table}. Moreover, we also plot the CMC curve in Fig. \ref {ablation figure} to show the detailed match rate results from top 1 to top 20 in the three test sets of VehicleID and VeRi. The notations $q=2$, $q=4$, or $q=8$ indicate that the number of branches in the stripe-based branch is set to 2, 4, or 8, respectively.
\subsubsection{Effectiveness of the Attribute-aware Branch}
To validate the effectiveness and superiority of the attribute-aware branch, we compare it with our baseline (i.e., ID cross-entropy loss based on ResNet-50), which is shown in the first row of Table \ref {ablation table}. As shown in Table \ref {ablation table}, although models with $L_{ID}$ have already reached a considerably higher accuracy compared with the results reported in \cite{zhou2018viewpoint, guo2019two}, the proposed attribute-aware branch, which includes $L_{ID}$ and $L_{Model}$, demonstrate further improvement. In comparison with the baseline with only $L_{ID}$, the improvement of the attribute-aware branch under the three test sets of VehicleID in terms of Rank-1 accuracy is 8.1\%, 6.8\%, and 5.7\%, respectively. On VeRi, the improvements in Rank-1 and mAP are 1.1\% and 5.7\%, respectively. The consistent performance improvement on both datasets confirms the superiority of the proposed attribute-aware branch. Thus, the correctness of adding discriminative vehicle attribute information to the Re-ID model for structured feature embedding learning is demonstrated.
\par
Furthermore, we conduct qualitative visualization experiments on the two branches of SAN to determine how the proposed two-branch stripe-based and attribute-aware network works. The feature maps after the first max-pooling layer in the backbone network are visualized because the feature maps of the following layers become increasingly abstract as the layer deepens. For the feature maps of the baseline, stripe-based branch, attribute-aware branch, and SAN, the most active channel in the feature map of SAN is selected, and the corresponding channel of baseline, stripe-based branch and attribute-aware branch are visualized. As shown in Fig. \ref {visualization}, the baseline model only focuses on the outline of the vehicle and ignores subtle details. Compared to the baseline model, the attribute-aware branch focuses more on the car-head region. In this branch, some subtle clues in the car-head region that can identify a vehicle (e.g., headlights, logo, and personalized decorations) are highly enhanced. Furthermore, this branch pays more attention on the global vehicle shape information, which is beneficial for vehicle model classification. The visualization results strongly prove the effectiveness of the proposed attribute-aware branch.
\subsubsection{Effectiveness of the Stripe-based Branch}
The next contribution needing to be investigated is the effectiveness of the stripe-based branch, which is realized by the horizontal average pooling and dimension-reduced convolutional layers. Thus, we conduct ablation experiments to evaluate the benefit of the stripe-based branch quantitatively by removing the attribute-aware branch from the SAN model. As shown in Fig. \ref{Fig1}, $q$ determines the granularity of the part-level features. Therefore, contrast experiments should be conducted to determine the number of $q$ that will produce the most improvement. As shown in Table \ref {ablation table}, as $q$ increases, the retrieval accuracy improves. When $q=8$, the accuracy reaches the highest point; hence, this value actually promotes the discriminative ability of the part-level features most.
\par
As shown in Table \ref {ablation table}, in comparison with the baseline with only $L_{ID}$, the stripe-based branch with $q=8$ exhibits 8.1\%, 8.2\%, and 11.2\% improvement in Rank-1 accuracy under the three test sets of VehicleID. On VeRi, the improvements in Rank-1 and mAP are 4.0\% and 9.1\%, respectively. 
The improvement in large and arduous test sets is larger than that in small and easy ones, which demonstrates the robustness and scalability of the stripe-based branch. This finding validates the effectiveness and essentiality of the stripe-based branch. Moreover, as Fig. \ref {visualization} illustrates, compared with the baseline model, the stripe-based branch focuses more on the windshield region, which contains nearly all the subtle yet discriminative traces (e.g., personalized decorations, inspection marks, and damage traces) that play a major role in vehicle Re-ID. The visualization result strongly proves the effectiveness of the proposed striped-based branch, as well as reveals the causes of the considerable improvements.
\par
 Finally, we combine the attribute-aware branch and stripe-based branch with $q=8$ together to get the best performance and evaluate it on both VehicleID and VeRi datasets. On VehicleID, the final Rank-1 accuracies on three test sets are 79.7\%, 78.4\%, 75.6\%, respectively; whereas on VeRi, the final mAP and Rank-1 accuracy are 72.5\% and 93.3\%, respectively. As shown in Fig. \ref {visualization}, the car-head and windshield regions are both activated in SAN. Thus, subtle yet discriminative traces in both regions are spotted by the SAN model. In conclusion, the proposed SAN provides evident benefits for vehicle Re-ID tasks.
\subsection{Comparison with State-of-the-art Methods}
\subsubsection{Comparison on VehicleID}
\begin{table}[]
\renewcommand{\arraystretch}{1.3}
\renewcommand{\multirowsetup}{\centering}
\caption{Comparison (\%) with State-of-the-art Methods on VehicleID}
\label{vehicleid table}
\centering
\begin{tabular}{c|cc|cc|cc}
\hlinew{1.5pt}
\multirow{2}{*}{Methods} & \multicolumn{2}{c|}{Test = 800} & \multicolumn{2}{c|}{Test = 1600} & \multicolumn{2}{c}{Test = 2400} \\
                        & r=1            & r=5            & r=1             & r=5            & r=1             & r=5            \\ \hlinew{1pt}
BOW-CN\cite{zheng2015scalable}                  & 13.14          & 22.69          & 12.94           & 21.09          & 10.20           & 17.89          \\
LOMO\cite{liao2015person}                    & 19.74          & 32.14          & 18.95           & 29.46          & 15.26           & 25.63          \\
GoogLeNet\cite{yang2015large}               & 47.90          & 67.43          & 43.45           & 63.53          & 38.24           & 59.51          \\
NuFACT\cite{liu2017provid}                  & 48.90          & 69.51          & 43.64           & 65.34          & 38.63           & 60.72          \\
FACT\cite{liu2016large}                    & 49.53          & 67.96          & 44.63           & 64.19          & 39.91           & 60.49          \\
DRDL\cite{liu2016deepdeep}                    & 49.0          & 73.5          & 42.8           & 66.8          & 38.2           & 61.6          \\
VAMI\cite{zhou2018viewpoint}                    & 63.12          & 83.25          & 52.87           & 75.12          & 47.34           & 70.29          \\
TAMR\cite{guo2019two}                    & 66.02          & 79.71          & 62.90           & 76.80          & 59.69           & 73.87          \\
QD-DLF\cite{zhu2019vehicle}                  & 72.32          & 92.48          & 70.66           & 88.90          & 64.14           & 83.37          \\
PN\cite{he2019part}        & 78.4           & 92.3           & 75.0            & 88.3           & 74.2            & 86.4           \\ \hlinew{1pt}
SAN (ours)               & \textbf{79.7}           & \textbf{94.3}           & \textbf{78.4}            & \textbf{91.3}           & \textbf{75.6}            & \textbf{88.3}           \\ \hlinew{1.5pt}
\end{tabular}
\end{table}
The performance comparison of the proposed SAN and several state-of-art methods under the VehicleID database is shown in Table \ref{vehicleid table}. It can be observed that the proposed SAN acquires the highest Rank-1 identification rates (i.e., 79.7\%, 78.4\%, and 75.6\% for Test800, Test1600, and Test2400, respectively) and Rank-5 identification rates (i.e., 94.3\%, 91.3\%, and 88.3\%, for Test800, Test1600, and Test2400, respectively) among the compared methods. The details of the analysis results are discussed below.
\par
Firstly, it can be found that deep learning-based methods (i.e., GoogLeNet \cite{yang2015large}, NuFACT \cite{liu2017provid}, FACT \cite{liu2016large}, DRDL \cite{liu2016deepdeep}, VAMI \cite{zhou2018viewpoint}, TAMR \cite{guo2019two}, QD-DLF \cite{zhu2019vehicle}, and PN \cite{he2019part}) are superior to traditional methods (i.e., BOW-CN \cite{zheng2015scalable} and LOMO \cite{liao2015person}) in the VehicleID database. 
\par
Secondly, the proposed SAN method outperforms all deep learning-based methods under comparison, including GoogLeNet \cite{yang2015large}, NuFACT \cite{liu2017provid}, FACT \cite{liu2016large}, DRDL \cite{liu2016deepdeep}, VAMI \cite{zhou2018viewpoint}, TAMR \cite{guo2019two}, QD-DLF \cite{zhu2019vehicle}, and PN \cite{he2019part}, on the three test sets under the VehicleID database. Specifically, PN \cite{he2019part}, which is the top deep learning-based vehicle Re-ID method, has reached 78.4\%, 75.0\%, and 74.2\% Rank-1 identification rates on Test800, Test1600, and Test2400, respectively, which are much higher than other deep learning-based vehicle Re-ID methods. However, these rates, as well as the Rank-5 identification rates, have been surpassed by the proposed SAN.
\subsubsection{Comparison on VeRi}
Table \ref{veri table} shows the comparison results of the proposed SAN and the existing state-of-art methods under the VeRi database.
It can be observed that the proposed SAN acquires the highest mAP (72.5\%), Rank-1 identification rate (93.3\%), and Rank-5 identification rate (97.1\%) among all methods. The details of the analysis results are discussed below.
\par
Firstly, in comparison with those handcrafted feature representation methods (i.e., LOMO \cite{liao2015person} and BOW-CN \cite{zheng2015scalable}), the proposed SAN method consistently defeats LOMO \cite{liao2015person} and BOW-CN \cite{zheng2015scalable} by higher mAP, Rank-1 identification rate and Rank-5 identification rate.
\begin{table}[]
\renewcommand{\arraystretch}{1.3}
\renewcommand{\multirowsetup}{\centering}
\caption{Comparison (\%) with State-of-the-art Methods on VeRi}
\label{veri table}
\centering
\begin{tabular}{c|ccc}
\hlinew{1.5pt}
Methods               & mAP   & r=1   & r=5   \\ \hlinew{1pt}
LOMO\cite{liao2015person}                  & 9.64  & 25.33 & 46.48 \\
BOW-CN\cite{zheng2015scalable}                & 12.20 & 33.91 & 53.69 \\
GoogLeNet\cite{yang2015large}              & 17.89 & 52.32 & 72.17 \\
FACT\cite{liu2016large}                   & 18.75 & 52.21 & 72.88 \\
DenseNet121\cite{huang2017densely}           & 45.06 & 80.27 & 91.12 \\
NuFACT\cite{liu2017provid}                & 48.47 & 76.76 & 91.42 \\
VAMI\cite{zhou2018viewpoint}                  & 50.13 & 77.03 & 90.82 \\
PROVID\cite{liu2017provid}                & 53.42 & 81.56 & 95.11 \\
Siamese-CNN+Path-LSTM\cite{shen2017learning} & 58.27 & 83.49 & 90.04 \\
QD-DLF\cite{zhu2019vehicle}                 & 61.83 & 88.50 & 94.46 \\ \hlinew{1pt}
SAN (ours)             & \textbf{72.5}  & \textbf{93.3}  & \textbf{97.1}  \\ \hlinew{1.5pt}
\end{tabular}
\end{table}
\par
Secondly, compared with those single-modal deep learning based vehicle Re-ID methods (i.e., GoogLeNet \cite{yang2015large}, FACT \cite{liu2016large}, DenseNet121 \cite{huang2017densely}, NuFACT \cite{liu2017provid}, VAMI \cite{zhou2018viewpoint}, and QD-DLF \cite{zhu2019vehicle}), the proposed SAN method demonstrates a higher accuracy improvement. Specifically, QD-DLF \cite{zhu2019vehicle}, which is the top single-modal deep learning-based vehicle Re-ID method, only obtains an mAP of 61.82\%, a Rank-1 ID rate of 88.50\%, and a Rank-5 ID rate of 94.46\%, which are considerably lower than those of the proposed SAN method. Moreover, it can be seen that VAMI \cite{zhou2018viewpoint} and QD-DLF \cite{zhu2019vehicle} do not obviously exhibit the superiority on the VeRi database, although they specially consider the discriminative part-level features.
\par
Thirdly, the proposed SAN method obviously defeats those multi-modal deep learning-based vehicle Re-ID methods (i.e., PROVID \cite{liu2017provid} and Siamese-CNN+Path-LSTM \cite{shen2017learning}) in terms of higher mAP, Rank-1 identification rate, and Rank-5 identification rate.
\section{Conclusion}
In this paper, we propose a two-branch SAN to learn efficient feature embedding for vehicle Re-ID. The SAN includes two branches, namely, stripe-based branch and attribute-aware branch. On the one hand, horizontal average pooling and dimension-reduced convolutional layers are inserted into the stripe-based branch to achieve part-level features. The attribute-aware branch, on the other hand, extract global feature under the supervision of vehicle attribute labels. The part-level and global features are then concatenated together to form the final descriptor of the input image for vehicle Re-ID. Using the deep learning feature learned by the proposed SAN, the adverse effect of the horizontal viewpoint variations is effectively resisted, and the vehicle Re-ID performance is greatly improved. Extensive component analysis and comparison experiments on VeRi and VehicleID databases demonstrate that the proposed method is superior to the existing state-of-the-art vehicle Re-ID methods.


%
\section*{Acknowledgment}
The authors would like to gratefully acknowledge the National Natural Science Foundation of China (Grant Nos. 61633019) and the Science Foundation of Chinese Aerospace Industry (Grant Nos. JCKY2018204B053).

\ifCLASSOPTIONcaptionsoff
  \newpage
\fi



%

\bibliographystyle{IEEEtran}      
\bibliography{IEEEexample}                        

\begin{thebibliography}{10}
\providecommand{\url}[1]{#1}
\csname url@samestyle\endcsname
\providecommand{\newblock}{\relax}
\providecommand{\bibinfo}[2]{#2}
\providecommand{\BIBentrySTDinterwordspacing}{\spaceskip=0pt\relax}
\providecommand{\BIBentryALTinterwordstretchfactor}{4}
\providecommand{\BIBentryALTinterwordspacing}{\spaceskip=\fontdimen2\font plus
\BIBentryALTinterwordstretchfactor\fontdimen3\font minus
  \fontdimen4\font\relax}
\providecommand{\BIBforeignlanguage}[2]{{%
\expandafter\ifx\csname l@#1\endcsname\relax
\typeout{** WARNING: IEEEtran.bst: No hyphenation pattern has been}%
\typeout{** loaded for the language `#1'. Using the pattern for}%
\typeout{** the default language instead.}%
\else
\language=\csname l@#1\endcsname
\fi
#2}}
\providecommand{\BIBdecl}{\relax}
\BIBdecl

\bibitem{liu2016deepdeep}
H.~Liu, Y.~Tian, Y.~Yang, L.~Pang, and T.~Huang, ``Deep relative distance
  learning: Tell the difference between similar vehicles,'' in
  \emph{Proceedings of the IEEE Conference on Computer Vision and Pattern
  Recognition}, 2016, pp. 2167--2175.

\bibitem{liu2016deep}
X.~Liu, W.~Liu, T.~Mei, and H.~Ma, ``A deep learning-based approach to
  progressive vehicle re-identification for urban surveillance,'' in
  \emph{European Conference on Computer Vision}.\hskip 1em plus 0.5em minus
  0.4em\relax Springer, 2016, pp. 869--884.

\bibitem{liu2016large}
X.~Liu, W.~Liu, H.~Ma, and H.~Fu, ``Large-scale vehicle re-identification in
  urban surveillance videos,'' in \emph{2016 IEEE International Conference on
  Multimedia and Expo (ICME)}.\hskip 1em plus 0.5em minus 0.4em\relax IEEE,
  2016, pp. 1--6.

\bibitem{zapletal2016vehicle}
D.~Zapletal and A.~Herout, ``Vehicle re-identification for automatic video
  traffic surveillance,'' in \emph{Proceedings of the IEEE Conference on
  Computer Vision and Pattern Recognition Workshops}, 2016, pp. 25--31.

\bibitem{yuan2017hard}
Y.~Yuan, K.~Yang, and C.~Zhang, ``Hard-aware deeply cascaded embedding,'' in
  \emph{Proceedings of the IEEE international conference on computer vision},
  2017, pp. 814--823.

\bibitem{yan2017exploiting}
K.~Yan, Y.~Tian, Y.~Wang, W.~Zeng, and T.~Huang, ``Exploiting multi-grain
  ranking constraints for precisely searching visually-similar vehicles,'' in
  \emph{Proceedings of the IEEE International Conference on Computer Vision},
  2017, pp. 562--570.

\bibitem{xu2017learning}
Q.~Xu, K.~Yan, and Y.~Tian, ``Learning a repression network for precise vehicle
  search,'' \emph{arXiv preprint arXiv:1708.02386}, 2017.

\bibitem{cheng2016person}
D.~Cheng, Y.~Gong, S.~Zhou, J.~Wang, and N.~Zheng, ``Person re-identification
  by multi-channel parts-based cnn with improved triplet loss function,'' in
  \emph{Proceedings of the iEEE conference on computer vision and pattern
  recognition}, 2016, pp. 1335--1344.

\bibitem{zhu2017part}
F.~Zhu, X.~Kong, L.~Zheng, H.~Fu, and Q.~Tian, ``Part-based deep hashing for
  large-scale person re-identification,'' \emph{IEEE Transactions on Image
  Processing}, vol.~26, no.~10, pp. 4806--4817, 2017.

\bibitem{schumann2017person}
A.~Schumann and R.~Stiefelhagen, ``Person re-identification by deep learning
  attribute-complementary information,'' in \emph{Proceedings of the IEEE
  Conference on Computer Vision and Pattern Recognition Workshops}, 2017, pp.
  20--28.

\bibitem{su2017multi}
C.~Su, F.~Yang, S.~Zhang, Q.~Tian, L.~S. Davis, and W.~Gao, ``Multi-task
  learning with low rank attribute embedding for multi-camera person
  re-identification,'' \emph{IEEE transactions on pattern analysis and machine
  intelligence}, vol.~40, no.~5, pp. 1167--1181, 2017.

\bibitem{almazan2018re}
J.~Almazan, B.~Gajic, N.~Murray, and D.~Larlus, ``Re-id done right: towards
  good practices for person re-identification,'' \emph{arXiv preprint
  arXiv:1801.05339}, 2018.

\bibitem{bai2018group}
Y.~Bai, Y.~Lou, F.~Gao, S.~Wang, Y.~Wu, and L.-Y. Duan, ``Group-sensitive
  triplet embedding for vehicle reidentification,'' \emph{IEEE Transactions on
  Multimedia}, vol.~20, no.~9, pp. 2385--2399, 2018.

\bibitem{kumar2019vehicle}
R.~Kumar, E.~Weill, F.~Aghdasi, and P.~Sriram, ``Vehicle re-identification: an
  efficient baseline using triplet embedding,'' \emph{arXiv preprint
  arXiv:1901.01015}, 2019.

\bibitem{shen2017learning}
Y.~Shen, T.~Xiao, H.~Li, S.~Yi, and X.~Wang, ``Learning deep neural networks
  for vehicle re-id with visual-spatio-temporal path proposals,'' in
  \emph{Proceedings of the IEEE International Conference on Computer Vision},
  2017, pp. 1900--1909.

\bibitem{zhou2018viewpoint}
Y.~Zhou, L.~Shao, and A.~Dhabi, ``Viewpoint-aware attentive multi-view
  inference for vehicle re-identification,'' in \emph{Proc. IEEE Conf. Comp.
  Vis. Patt. Recogn}, vol.~2, 2018.

\bibitem{he2019part}
B.~He, J.~Li, Y.~Zhao, and Y.~Tian, ``Part-regularized near-duplicate vehicle
  re-identification,'' in \emph{Proceedings of the IEEE Conference on Computer
  Vision and Pattern Recognition}, 2019, pp. 3997--4005.

\bibitem{Wang2017Orientation}
Z.~Wang, L.~Tang, X.~Liu, Z.~Yao, and X.~Wang, ``Orientation invariant feature
  embedding and spatial temporal regularization for vehicle
  re-identification,'' in \emph{IEEE International Conference on Computer
  Vision}, 2017.

\bibitem{khorramshahi2019attention}
P.~Khorramshahi, N.~Peri, A.~Kumar, A.~Shah, and R.~Chellappa, ``Attention
  driven vehicle re-identification and unsupervised anomaly detection for
  traffic understanding,'' in \emph{Proceedings of the IEEE Conference on
  Computer Vision and Pattern Recognition Workshops}, 2019, pp. 239--246.

\bibitem{khorramshahi2019dual}
P.~Khorramshahi, A.~Kumar, N.~Peri, S.~S. Rambhatla, J.-C. Chen, and
  R.~Chellappa, ``A dual path modelwith adaptive attention for vehicle
  re-identification,'' \emph{arXiv preprint arXiv:1905.03397}, 2019.

\bibitem{wang2019vehicle}
P.~Wang, B.~Jiao, L.~Yang, Y.~Yang, S.~Zhang, W.~Wei, and Y.~Zhang, ``Vehicle
  re-identification in aerial imagery: Dataset and approach,'' \emph{arXiv
  preprint arXiv:1904.01400}, 2019.

\bibitem{guo2019two}
H.~Guo, K.~Zhu, M.~Tang, and J.~Wang, ``Two-level attention network with
  multi-grain ranking loss for vehicle re-identification,'' \emph{IEEE
  Transactions on Image Processing}, 2019.

\bibitem{teng2018scan}
S.~Teng, X.~Liu, S.~Zhang, and Q.~Huang, ``Scan: Spatial and channel attention
  network for vehicle re-identification,'' in \emph{Pacific Rim Conference on
  Multimedia}.\hskip 1em plus 0.5em minus 0.4em\relax Springer, 2018, pp.
  350--361.

\bibitem{zhu2019vehicle}
J.~Zhu, H.~Zeng, J.~Huang, S.~Liao, Z.~Lei, C.~Cai, and L.~Zheng, ``Vehicle
  re-identification using quadruple directional deep learning features,''
  \emph{IEEE Transactions on Intelligent Transportation Systems}, 2019.

\bibitem{li2017deep}
Y.~Li, Y.~Li, H.~Yan, and J.~Liu, ``Deep joint discriminative learning for
  vehicle re-identification and retrieval,'' in \emph{2017 IEEE International
  Conference on Image Processing (ICIP)}.\hskip 1em plus 0.5em minus
  0.4em\relax IEEE, 2017, pp. 395--399.

\bibitem{hou2019multi}
J.~Hou, H.~Zeng, L.~Cai, J.~Zhu, J.~Chen, and K.-K. Ma, ``Multi-label learning
  with multi-label smoothing regularization for vehicle re-identification,''
  \emph{Neurocomputing}, vol. 345, pp. 15--22, 2019.

\bibitem{zheng2019attributes}
A.~Zheng, X.~Lin, C.~Li, R.~He, and J.~Tang, ``Attributes guided feature
  learning for vehicle re-identification,'' \emph{arXiv preprint
  arXiv:1905.08997}, 2019.

\bibitem{zhao2019structural}
Y.~Zhao, C.~Shen, H.~Wang, and S.~Chen, ``Structural analysis of attributes for
  vehicle re-identification and retrieval,'' \emph{IEEE Transactions on
  Intelligent Transportation Systems}, 2019.

\bibitem{nguyen2019vehicle}
K.-T. Nguyen, T.-H. Hoang, M.-T. Tran, T.-N. Le, N.-M. Bui, T.-L. Do, V.-K.
  Vo-Ho, Q.-A. Luong, M.-K. Tran, T.-A. Nguyen \emph{et~al.}, ``Vehicle
  re-identification with learned representation and spatial verification and
  abnormality detection with multi-adaptive vehicle detectors for traffic video
  analysis,'' in \emph{Proceedings of the IEEE Conference on Computer Vision
  and Pattern Recognition Workshops}, 2019, pp. 363--372.

\bibitem{huang2019multi}
T.-W. Huang, J.~Cai, H.~Yang, H.-M. Hsu, and J.-N. Hwang, ``Multi-view vehicle
  re-identification using temporal attention model and metadata re-ranking,''
  in \emph{AI City Challenge Workshop, IEEE/CVF Computer Vision and Pattern
  Recognition (CVPR) Conference, Long Beach, California}, 2019.

\bibitem{chatfield2014return}
K.~Chatfield, K.~Simonyan, A.~Vedaldi, and A.~Zisserman, ``Return of the devil
  in the details: Delving deep into convolutional nets,'' \emph{arXiv preprint
  arXiv:1405.3531}, 2014.

\bibitem{szegedy2017inception}
C.~Szegedy, S.~Ioffe, V.~Vanhoucke, and A.~A. Alemi, ``Inception-v4,
  inception-resnet and the impact of residual connections on learning,'' in
  \emph{Thirty-First AAAI Conference on Artificial Intelligence}, 2017.

\bibitem{he2016deep}
K.~He, X.~Zhang, S.~Ren, and J.~Sun, ``Deep residual learning for image
  recognition,'' in \emph{Proceedings of the IEEE conference on computer vision
  and pattern recognition}, 2016, pp. 770--778.

\bibitem{sun2018beyond}
Y.~Sun, L.~Zheng, Y.~Yang, Q.~Tian, and S.~Wang, ``Beyond part models: Person
  retrieval with refined part pooling (and a strong convolutional baseline),''
  in \emph{Proceedings of the European Conference on Computer Vision (ECCV)},
  2018, pp. 480--496.

\bibitem{luo2019bag}
H.~Luo, Y.~Gu, X.~Liao, S.~Lai, and W.~Jiang, ``Bag of tricks and a strong
  baseline for deep person re-identification,'' in \emph{Proceedings of the
  IEEE Conference on Computer Vision and Pattern Recognition Workshops}, 2019,
  pp. 0--0.

\bibitem{chen2019partition}
H.~Chen, B.~Lagadec, and F.~Bremond, ``Partition and reunion: A two-branch
  neural network for vehicle re-identification,'' in \emph{Proceedings of the
  IEEE Conference on Computer Vision and Pattern Recognition Workshops}, 2019,
  pp. 184--192.

\bibitem{huang2017densely}
G.~Huang, Z.~Liu, L.~Van Der~Maaten, and K.~Q. Weinberger, ``Densely connected
  convolutional networks,'' in \emph{Proceedings of the IEEE conference on
  computer vision and pattern recognition}, 2017, pp. 4700--4708.

\bibitem{zheng2015scalable}
L.~Zheng, L.~Shen, L.~Tian, S.~Wang, J.~Wang, and Q.~Tian, ``Scalable person
  re-identification: A benchmark,'' in \emph{Proceedings of the IEEE
  international conference on computer vision}, 2015, pp. 1116--1124.

\bibitem{liao2015person}
S.~Liao, Y.~Hu, X.~Zhu, and S.~Z. Li, ``Person re-identification by local
  maximal occurrence representation and metric learning,'' in \emph{Proceedings
  of the IEEE conference on computer vision and pattern recognition}, 2015, pp.
  2197--2206.

\bibitem{yang2015large}
L.~Yang, P.~Luo, C.~Change~Loy, and X.~Tang, ``A large-scale car dataset for
  fine-grained categorization and verification,'' in \emph{Proceedings of the
  IEEE Conference on Computer Vision and Pattern Recognition}, 2015, pp.
  3973--3981.

\bibitem{liu2017provid}
X.~Liu, W.~Liu, T.~Mei, and H.~Ma, ``Provid: Progressive and multimodal vehicle
  reidentification for large-scale urban surveillance,'' \emph{IEEE
  Transactions on Multimedia}, vol.~20, no.~3, pp. 645--658, 2017.

\end{thebibliography}

%








\end{document}